\documentclass[fleqn,usenatbib,usedcolumn]{mnras}

\usepackage{newtxtext,newtxmath}
\usepackage{booktabs}
\usepackage{tikz}
\usetikzlibrary{arrows}
\usepackage{color}
\usepackage{graphicx}
\usepackage{caption}
\usepackage{subcaption}
\usepackage{multirow}
\usetikzlibrary{backgrounds}
\usepackage{threeparttable}
\usepackage{svg}
\usepackage{pdflscape}
\DeclareRobustCommand{\VAN}[3]{#2}
\let\VANthebibliography\thebibliography
\def\thebibliography{\DeclareRobustCommand{\VAN}[3]{##3}\VANthebibliography}
\usepackage{graphicx}	
\usepackage{amsmath}	

\onecolumn

\title[Short title, max. 45 characters]{ A self-regulated convolutional neural network for classifying variable stars}

\author[]{
\large
F. P\'erez-Galarce$^{1,9},$ \text{  }
J. Martínez-Palomera$^{2,3},$ \text{  }
 K. Pichara$^{1}$,  \text{  } P. Huijse$^{4,5,6}$, \text{  } M. Catelan$^{6,7,8}$  \text{  }
\\
$^{1}$ Department of Computer Science, School of Engineering, Pontificia Universidad Católica de Chile, 7820436 Santiago, Chile \\
$^{2}$ University of Maryland Baltimore County, Baltimore, MD, USA.\\
$^{3}$ NASA Goddard Space Flight Center, Greenbelt, MD, USA\\
$^{4}$ Institute of Astronomy, KU Leuven, Celestijnenlaan 200D, B-3001 Leuven, Belgium\\
$^{5}$ Instituto de Informática, Universidad Austral de Chile, Valdivia, Chile\\
$^{6}$ Millennium Institute of Astrophysics, Nuncio Monseñor Sotero Sanz 100, Of. 104, Providencia, Santiago, Chile\\
$^{7}$  Instituto de Astrofísica, Pontificia Universidad Católica de Chile, Av. Vicuña Mackenna 4860, 7820436 Macul, Santiago, Chile\\
$^{8}$ Centro de Astro-Ingeniería, Pontificia Universidad Católica de Chile, Av. Vicuña Mackenna 4860, 7820436 Macul, Santiago, Chile\\
$^{9}$ \color{black} Facultad de Ingeniería y Negocios, Universidad de Las Américas, Sede Providencia,
Manuel Montt 948, Santiago, Chile\\
}

\date{Accepted XXX. Received YYY; in original form ZZZ}

\pubyear{2015}

\begin{document}
\label{firstpage}
\pagerange{\pageref{firstpage}--\pageref{lastpage}}
\maketitle
\begin{abstract}

Over the last two decades, machine learning models have been widely applied and have proven effective in classifying variable stars, particularly with the adoption of deep learning architectures such as convolutional neural networks, recurrent neural networks, and transformer models. While these models have achieved high accuracy, they require high-quality, representative data and a large number of labelled samples for each star type to generalise well, which can be challenging in time-domain surveys. This challenge often leads to models learning and reinforcing biases inherent in the training data, an issue that is not easily detectable when validation is performed on subsamples from the same catalogue. The problem of biases in variable star data has been largely overlooked, and a definitive solution has yet to be established. In this paper, we propose a new approach to improve the reliability of classifiers in variable star classification by introducing a self-regulated training process. This process utilises synthetic samples generated by a physics-enhanced latent space variational autoencoder, incorporating six physical parameters from {\em Gaia} Data Release 3. Our method features a dynamic interaction between a classifier and a generative model, where the generative model produces ad-hoc synthetic light curves to reduce confusion during classifier training and populate underrepresented regions in the physical parameter space.   \color{black} Experiments conducted under various scenarios demonstrate that our self-regulated training approach outperforms traditional training methods for classifying variable stars on biased datasets, showing statistically significant improvements. \color{black}
\end{abstract}
 



\twocolumn
\section{Introduction}

Over recent decades, machine learning (ML) models have become a key component in the analysis of vast astronomical time-series surveys, providing an automated solution to the complex task of pattern recognition within extensive datasets. Variable stars, in particular, are of interest for their critical role in astronomical measuring, especially types such as RR Lyrae and Cepheids that are key to determining cosmic distances \citep{beaton2016carnegie}. Consequently, there has been a massive motivation to refine ML strategies to enhance the precision of variable star classification, which currently relies heavily on deep learning (DL) models \citep{aguirre2018deep, naul2018recurrent, becker2020scalable, zhang2021classification, donoso2022astromer, perez2023informative}.

However, applying automatic ML-based classification of variable stars in real-world scenarios presents significant challenges. One major hurdle is the creation of labelled datasets that are both massive and unbiased. Moreover, certain characteristics of variable star classes, such as class overlapping and class imbalance, make classification particularly difficult. Imbalanced classes and data shift issues are well-documented challenges in variable stars datasets \citep{debosscher2009automated, aguirre2018deep,burhanudin2021light,garcia2022improving,perez2021informative, perez2023informative}. 

The data shift problem comes in various shapes, such as covariate shift, target shift, and conditional shift \citep{candela2009dataset}. The covariate shift assumes that the features in the testing set (or target set), such as period or metallicity, have a different joint density distribution compared to the training set (or source set), even when assuming a representative conditional distribution. In other words, \(p^S(\mathbf{x}) \neq p^T(\mathbf{x})\), and \(p^T(y|\mathbf{x}) = p^S(y|\mathbf{x})\), where \(p^S(\mathbf{x})\) and \(p^T(\mathbf{x})\) are the probability density distributions of features in the source and target sets, respectively. Additionally, \(p^T(y|\mathbf{x}) \text{ and } p^S(y|\mathbf{x})\) represents the conditional distributions of the label given the features. A target shift indicates a mismatch in the distribution of label proportions of each class; it considers \(p^T(y) \neq p^S(y)\), but \(p^S(\mathbf{x}|y) = p^T(\mathbf{x}|y)\). The conditional shift occurs when the relationship between one or all classes and the features changes; for example, for identical feature values, a star type might have a different probability value in the testing set compared to the training set. This data shift is described by \(p^S(y|\mathbf{x}) \neq p^T(y|\mathbf{x})\) and \(p^T(y) = p^S(y)\). 

Data shifts in variable star classification problems may arise due to differences in the technical specifications of the instruments, the scheduling of the observations or the criteria used by the experts who analyse and label the light curves. One such bias emerges during the labelling process, where astronomers may more frequently identify specific types of stars that are easier to classify. This bias in the astronomical labelling process is detailed in \citep{cabrera2014systematic}. Another source of bias arises from the technical limitations of the telescopes, particularly the detection and characterisation of faint objects due to distance or less luminous \citep{richards2012overcoming}. Observation scheduling of telescopes can also induce data shift. Selection biases emerge when specific sky regions are preferentially observed, potentially leading to covariate shifts if these regions differ from others in critical characteristics. Variations in observation times, cadence, and environmental conditions can also alter the data collected, affecting the distribution and types of celestial objects observed. Furthermore, updates to telescope technology or changes in instrumentation can result in data that are different in sensitivity or resolution, leading to concept shifts where the relationship between inputs and outputs changes; this issue is also referred to as domain adaptation (or transfer learning). It has been thoroughly explored in the context of variability surveys in \citep{benavente2017automatic}.  The aforementioned biases are even more challenging when the number of observations per object is small, or the signal-to-noise ratio is low, as the underlying patterns of each light curve become more difficult for end-to-end neural networks to identify.

Imbalanced data problem refers to scenarios where the number of instances in each class within a dataset significantly differs. This can severely affect the performance of ML algorithms, which typically assume an equal number of examples for each class to function optimally.  In variability surveys, for instance, some variable stars might be much rarer than others; e.g., type II Cepheids are less frequent than RR Lyrae stars \citep{catelan2014pulsating}. This issue can lead to miscalibrated models for predicting the more common classes. Addressing imbalanced datasets often requires specialised techniques such as data augmentation \citep{aguirre2018deep,hosenie2020imbalance} or applying weighting loss functions \citep{burhanudin2021light} during model training to ensure that rare classes are treated adequately. 

Significant challenges persist in mitigating the impact of data shift despite various strategies such as domain adaptation, regular retraining, feature engineering, instance weighting, ad-hoc loss functions, and data augmentation. These approaches include high computational costs, the complexity of tuning models to new data without overfitting, and the difficulty of obtaining timely and representative data for retraining. These limitations hinder their practical application and scalability in real-world settings. Generative artificial intelligence, which is responsible for recent breakthroughs in science and technology across various fields, has emerged as a scalable and practical approach for generating realistic light curves \citep{martinez2022deep, garcia2022improving}, offering a promising solution to mitigate biases and address class imbalances. However, this method requires careful application, as it can introduce additional biases into the dataset and may be challenging to scale if focused solely on balancing the number of light curve cases. Therefore, ad-hoc training algorithms are essential to make fruitful use of these new models.

 This paper proposes a novel approach to train more reliable variable star classifiers by leveraging recent advancements in synthetic data generation based on DL models. We propose integrating a generative model with a classifier in a cooperative framework; our approach dynamically enhances the learning process with synthetic examples in underrepresented areas. During the classifier training, the synthetic samples, which are focused on mitigating biases and imbalance problems, are initially obtained from the stellar physical parameter space, including effective temperature, period, metallicity, absolute magnitude, surface gravity, and radius. These samples are then processed by a trained physics-enhanced latent space variational autoencoder (PELS-VAE), which returns a synthetic light curve. We highlight that sampling from the physical parameter space, which is a low-dimensional space, allows us to manage the over/under-represented zones when generating new light curves. Our method adjusts the classifier training trajectory through the injection of new objects from the generative model. We propose five policies for defining the number of samples for each class, some aimed at populating classes where the confusion is most significant according to the confusion matrix in the current training epoch. A mask-based training scheme is included to avoid competition between real and synthetic data.  We also provide a set of experiments, considering two types of data shift, that assess the classifier performance under variations in the signal-to-noise ratio and the sequence length, highlighting where synthetic samples are most relevant for reducing data shift and the impact of class imbalance. Finally, we demonstrate that our synthetic light curves can assist in training more reliable classifiers and optimising hyperparameters, which remains a challenging task in current DL-based architectures.

The organisation of our paper is as follows: Section \ref{relatedwork} reviews the existing methodologies in variable star classifiers. Section \ref{Method} delineates our self-regulated convolutional neural network methodology. Section \ref{Data} expounds on the dataset, incorporating physical parameters from the {\em Gaia} Data Release 3 \citep[DR3;][]{creevey2023gaia}  and light curves obtained from the Optical Gravitational Lensing Experiment survey III \citep[OGLE III;][]{udalski2008optical}. Section \ref{Results} details our set of experiments and findings, and Section \ref{conclusion} provides a conclusion, summarising the principal discoveries and suggesting avenues for future work.

\section{Related work}
\label{relatedwork}

\subsection{Overcoming biases in variable stars classifiers}

One decade ago, \cite{richards2011active} brought attention to the prevalent issue of sample selection bias. They advocate for active learning, co-training, self-training and importance-weighting cross-validation techniques to face this challenge. Active learning, a framework where the classifier selectively queries the most informative objects for labelling, enhanced classifier performance, surpassing other approaches in terms of classification accuracy using data from the All-Sky Automated Survey for Supernovae \citep[ASAS-SN;][]{jayasinghe2019asas}. However, despite its effectiveness, traditional active learning is notably time-intensive as it requires ongoing human involvement to identify and label the most informative instances. 

\cite{vilalta2013machine} addressed the issue of dataset shift in classifying Cepheid variable stars from galaxies at different distances. They introduced a novel two-step approach to align the distributions of period and apparent magnitude features between source and target datasets, enhancing coherence without modifying the trained model or re-weighting samples. This adjustment is estimated using maximum log-likelihood with respect to the probability density of the training data.  Once the datasets are aligned, traditional classifiers are employed, including neural networks (NNs), support vector machines (SVM), random forest (RF), and decision trees. This method results in classification accuracy comparable to scenarios where labels are available for the target data.

\cite{benavente2017automatic} proposed a probabilistic framework for a survey-invariant variable star classification, addressing the domain adaptation challenge. The framework utilises Gaussian mixture models to represent the feature distributions of both the source and target surveys. This unsupervised model facilitates the invariant transformation of feature distributions from source to target surveys, including translation, rotation, and scaling of each Gaussian component. The effectiveness of the model was assessed through classifications in three catalogues: \textit{Experience de Recherche d'Objets Sombres II}  \citep[EROS;][]{tisserand2007limits}, the MAssive Compact Halo Object \citep[MACHO;][]{alcock1997macho}, and the High Cadence Transient Survey \citep[HiTS;][]{forster2016high}. RF and SVM classifiers were trained to evaluate domain adaptation performance. Their results show that transforming the features significantly improves classification accuracy. 

\cite{perez2021informative}  contributed with an informative Bayesian model selection technique for RR Lyrae stars that incorporates survey-invariant physical characteristics, such as the period of variable stars, into the assessment stage. This approach is based on a modified marginal likelihood estimation, where priors are incorporated through data manipulation based on deterministic ranges for characteristic descriptors. This strategy outperforms standard techniques for selecting classifiers, even when models include penalization. They demonstrated that even straightforward information from descriptor ranges can help to select better classifiers. Despite this, scalability remains an issue highlighted in this contribution, as the marginal likelihood estimation requires time-consuming sampling methods.

\cite{burhanudin2021light} presented a recurrent neural network (RNN) focused on handling the imbalanced problem observed in the Gravitational-wave Optical Transient Observer survey \cite[GOTO;][]{steeghs2022gravitational}. The proposed RNN consider two types of inputs; on the one hand, they include light curve information, which considers time, magnitude and photometric error; on the other hand, they use contextual information, namely, the Galactic coordinates expressed in degrees and the distance measured in arc-seconds to the closest galaxy listed in the Galaxy List for the Advanced Detector Era catalogue \cite[GLADE;][]{dalya2018glade}. The dataset contains variable stars, active galactic nuclei and supernovas, having 99 per cent of objects belonging to the majority class (variable stars). They highlight the crucial role played by loss functions, where a focal loss is proposed \citep{lin2017focal}, and the contextual information, suggesting that the strength of a classification system lies not just in the structure of the model but equally in the approach to data handling.

\cite{perez2023informative} introduced a method that facilitates the integration of expert knowledge during the training phase of NNs. This approach enhances classifier reliability through a novel regularisation technique. This regularisation involves incorporating expert knowledge into ML models via an interval representation of characteristic features. A training scheme with two masks is designed to inject this knowledge, where one mask is focused on learning from training data and the other is focused on learning from expert knowledge, the latter acting as a regularisation. Although this method surpassed baseline models in performance, the required computation of features may not be scalable for streaming pipelines, highlighting a gap between this methodological advance and its practical applications. 

To summarise, variable star classifiers have a latent drawback related to the data shift problem in training sets. This issue has been faced a few times, but no definite solution (scalable and reliable for online classifiers) has been established. The class imbalance problem, which is also very relevant in variable stars, has been studied separately. New methodological proposals should tackle both problems jointly to provide transferable classifiers. From the reviewed papers, we can highlight that treating the data shift problem by improving data or creating synthetic data is a promising path to training more reliable classifiers.  

\subsection{Deep learning-based classifiers}
During the last decade, variable star classification has been transformed by DL models; these advances have been motivated by the vast amount of data from forthcoming projects such as the Vera C. Rubin Observatory's Legacy Survey of Space and Time  \cite[LSST;][]{ivezic2008large}, where Terabytes of new information must be processed each night. These new technological advances have led to the development of end-to-end ML models that avoid computing descriptors or at least reduce the number of descriptors to essentials (e.g. period).

\cite{mahabal2017deep} offered an application of convolutional neural networks (CNNs) for classifying light curves from the Catalina Real-Time Transient Survey (CRTS; \cite{drake2009first}). In their method, light curves are converted into two-dimensional images, termed $\Delta m$$\Delta t$-images, characterising the change in magnitude ($\Delta m$) across varying time intervals ($\Delta t$); these $\Delta m$$\Delta t$-images are inputs for CNNs to classify. They achieved an 83\% accuracy rate, estimated by a five-random train-test split approach, in classifying periodic variable stars into seven classes, obtaining a result comparable with the performance of RF using manually crafted features. 

\cite{naul2018recurrent} developed a recurrent autoencoder that incorporates measurement errors into the loss function to weigh the observations. The representations learned by this autoencoder outperformed conventional methods based on manual feature construction when training traditional classifiers such as RFs. This direction of using artificial NNs, both recurrent and convolutional, for classifying astronomical objects has frequently matched or exceeded the predictive accuracy of RFs using handcrafted features \citep{carrasco2018deep, aguirre2018deep, becker2020scalable}. \cite{naul2018recurrent} conducted experiments using ASAS, LIncoln Near-Earth Asteroid Research \citep[LINEAR;][]{sesar2013exploring}, and MACHO surveys, obtaining mean accuracy of 98.77\%, 97.10\%, and 93.59\%, respectively, using a traditional cross-validation approach. 

\cite{aguirre2018deep} presented a CNN architecture for classifying variable stars using light curves from multiple astronomical surveys. The model uses a more straightforward representation of differences in time and magnitudes in consecutive observations,  namely the $(\Delta t_i, \Delta m_i)_{i=1}^L$ representation. The CNN was evaluated using data from the OGLE-III, VISTA Variables in the Vía Láctea ESO public \citep[VVV;][]{minniti2010vista}, and Convection, Rotation and planetary Transit \citep[CoRoT;][]{auvergne2009corot} surveys, each characterised by unique filters, cadences, and sky coverage. It achieved classification accuracy comparable to or superior to an RF classifier.  
A data augmentation technique was employed to balance the training data across classes. Capable of classifying both classes and subclasses, the model reached 85\% accuracy on the validation set. These findings underscore the potential of input representation and CNNs for efficient, large-scale classification of variable stars.

\cite{tsang2019deep} provided a model to learn light curve representations and classify objects with novel patterns. To achieve this, they trained an autoencoder and a classifier together. The model was trained using ASAS data. The pre-processing and training approach was similar to that of \cite{naul2018recurrent}, using 200 observations to train the autoencoder, with $\Delta$-time and normalised magnitude as the input representation. The loss function was defined by $\mathcal{L}_{\rm AE} + (\lambda \mathcal{L}_{\rm GMM} + \mathcal{L}_{\rm CE})$, incorporating the autoencoder loss, Gaussian mixture model loss, and cross-entropy loss, respectively. These loss components compete during the training process. The Gaussian mixtures, learned in the latent space, aim to detect outliers, i.e., new variability patterns. Experiments with sequential and joint training strategies showed similar results, confirming the feasibility of training these modules separately.

\cite{becker2020scalable}  proposed an end-to-end RNN approach for variable star classification using an input defined by differences in magnitude and time as \cite{aguirre2018deep}. The network is trained on three datasets: OGLE-III, {\em Gaia} and the Wide-field Infrared Survey Explorer \cite[WISE;][]{wright2010wide}. The results show that the network achieves classification accuracy comparable to that of the RF classifier while being faster and more scalable. 

\cite{jamal2020neural}  explored a variety of NN architectures, such as long-short term memory (LSTM), gated recurrent units (GRUs), temporal convolutional NNs (tCNNs) and dilated temporal CNNs. They compared two approaches: direct classifier networks and composite networks integrating autoencoder components. In addition, a significant improvement in classification performance is observed when metadata features are combined with the light curves, obtaining an accuracy increase of up to 20\%. Incorporating autoencoder components in the composite networks contributes to generating more distinguishable latent representations of the data, which correlates with the observed improvement in classification accuracy. This research declares that while RNNs and CNNs demonstrate aptitude in variable star classification, CNNs are distinguished by their reduced training time and memory resource requirements.

\cite{bassi2021classification} introduced two DL methods for classifying variable stars: a two-dimensional CNN (2D CNN) and the one-dimensional CNN-Long Short-Term Memory (1D CNN-LSTM). OGLE-III and CRTS were used for evaluation purposes. The 2D CNN method, which requires pre-processing the same as \cite{mahabal2017deep}, shows good performance when applied to the OGLE dataset, achieving an F$_1$ score of 0.85. F$_1$ score is a popular metric, defined as the harmonic mean of two more basic metrics: precision and recall, where precision is the proportion of true positives among all predicted positives, and recall is the proportion of true positives among all actual positives. However, its effectiveness declines with the CRTS dataset, where the F$_1$ score is 0.54. On the other hand, the 1D CNN-LSTM model, which directly utilises the original light curves without any preliminary modification,  obtains an F$_1$ value that is somewhat lower than the 2D CNN one (0.71 in OGLE and 0.49 in CRTS); it benefits from having fewer parameters and reduced computational demands. The authors proposed that future research could focus on refining the hyperparameters of the 1D CNN-LSTM model. Overall, this research highlights the effectiveness of the 1D CNN-LSTM in classifying light curves without requiring data preprocessing.

\cite{zhang2021classification} developed a novel NN called the cyclic-permutation invariant network tailored for including in architecture the periodic nature of phased light curves in variable star classification. This network addresses a fundamental limitation in traditional ANNs: their sensitivity to the initial phase in period-folded light curves, a factor that ideally should not influence classification outcomes. The proposed networks are designed to remain unaffected by phase shifts, ensuring more reliable classification. These networks were evaluated using datasets from three variable star surveys: ASAS-SN, MACHO, and OGLE-III. The invariant networks outperformed traditional non-invariant models and RNNs, achieving error rate reductions ranging from 4\% to 22\%. On the OGLE-III dataset, the invariant networks attained average per-class accuracy of 93.4\% and 93.3\%, significantly higher than the 70.5\% and 89.5\% accuracy achieved by RNN and RF models, respectively. Although the networks demonstrate effective performance on light curves with limited data points, their efficiency in a streaming context still needs to be tested, mainly because of their dependence on the period estimation accuracy.

\cite{abdollahi2023hierarchical} designed a hierarchical approach utilising deep CNNs for classifying variable stars. The proposed scheme operates in a two-tiered manner: initially determining the main type of a variable star, followed by sub type classification. This approach is particularly effective in enhancing accuracy for less-represented classes in the dataset. The methodology involves using light curves and the period of the stars as inputs for the ANNs. An essential step in data pre-processing includes folding light curves based on the period and their subsequent binning to achieve a uniform length. Light curves from OGLE-IV \citep{udalski2015ogle} are used, considering seven classes and sixteen subclasses. The implemented CNN model demonstrates impressive performance, with an accuracy of 98\% in main type classification and 93\% in sub type classification, outperforming RNN models in accuracy and training efficiency. 

\cite{kang2023periodic} introduced a novel DL  model for classifying periodic variable stars, addressing the challenge of imbalanced datasets. Their approach is based on an ensemble augmentation strategy, which involves creating synthetic light curves through Gaussian processes to strengthen underrepresented classes. The researchers designed two NN architectures: a multi-input RNN and a combined RNN-CNN structure. The multi-input RNN is fed by light curve information, period, and amplitude, whereas the RNN-CNN hybrid also incorporates light curve images into the CNN component. Comparative analysis revealed that the compound RNN-CNN model exhibited superior performance, achieving a macro F$_1$ of 0.75, outperforming the 0.71 scores achieved by the multi-input RNN model. It is worth noting that even providing the period as input, the macro F$_1$ did not exceed 0.75.

In summary, although there have been significant advancements in DL for classifying variable stars using light curves, most of them rely on period estimation, which can be time-consuming and prone to errors; our proposed classifier does not require period estimation. \color{black} In addition, the reported metrics demonstrate satisfactory results; however, they do not consider the data shift problem, hence, there is a high risk of overoptimism. \color{black} Thus, despite successes with complex architectures like CNNs and RNNs, a key challenge is their tendency to overfit, especially in biased datasets. This represents a pressing need for more reliable methods to automatically mitigate biases in astronomical datasets, ensuring classification models' robustness and generalisation capabilities to be deployed in ongoing or forthcoming telescopes.

\subsection{Generative models for variable stars}

The generation of synthetic data is a prevalent method to enhance model quality. This data can be produced through simulation methods based on analytical models or by applying domain knowledge to modify existing data, known as data augmentation. Simulation, influenced by theory-driven models, often requires substantial computational resources. Conversely, the domain of data augmentation has seen limited exploration in the literature \citep{castro2017uncertain, aguirre2018deep,kang2023periodic}. In this field, various constraints hinder these methods, particularly the sequential nature of the data. These methods face challenges in interpolating physical parameters and handling the irregular intervals between observations, presenting significant challenges to effective data enhancement.

A few papers have recently proposed deep generative modelling to design procedures based on generative adversarial networks (GANs) and variational autoencoders (VAEs) to create synthetic light curves of periodic stars. These approaches offer a novel alternative to address the data shift problem discussed in this paper.

\cite{garcia2022improving} proposed GANs for creating synthetic astronomical light curves to improve the classification of variable stars, addressing challenges like limited data sizes and class imbalance. They proposed a GAN-driven data augmentation strategy for conditional generation, considering the star type and physical characteristics of irregularly spaced time series. They introduced a new evaluation metric for selecting suitable GAN models and a resampling method designed to delay GAN overfitting during training. This resampling technique modifies the class distribution by controlling the probability of sampling from each class in a scheme without replacement, adaptable to highly imbalanced or balanced sets by adjusting a parameter \(\gamma\). Additionally, they developed two data augmentation techniques that generate plausible time series while preserving the properties of the original data, though these did not surpass the GAN-based methods.

\cite{martinez2022deep} proposed a deep generative model, a PELS-VAE, using a VAE to create synthetic light curves of periodic variable stars based on their labels and physical parameters. This model integrates a conditional VAE with alternative layers: temporal convolutional networks and RNNs (LSTM and GRU). Trained on OGLE-III dataset light curves and physical parameters from  {\em Gaia} DR2, it aims for a more accurate latent space representation. The model generates unseen light curves by specifying physical parameters, mapping these to the latent space through regression, and then sampling and decoding the latent vector. The transformation of RR Lyrae light curves from saw-tooth to sinusoidal shapes with temperature increase showcases the model's effectiveness. This research presents an ad-hoc generative model that proficiently produces realistic light curves for periodic variable stars based on their physical parameters, offering new directions for data augmentation methodologies research.

\cite{ding2024detection} proposed an unsupervised approach to classify contact binaries candidates; an autoencoder model is trained using synthetic light curves according to physical parameters, e.g., the mass ratio, the orbital inclination, the effective temperature of the primary star, the effective temperature of the secondary star. These parameters are provided to PHysics Of Eclipsing BinariEs software  \citep[PHOEBE;][]{prvsa2005computational} to create synthetic samples. Once the autoencoder is trained, a sequence of criteria is created to detect potential contact binaries.

According to the reviewed articles, the generation of synthetic samples employing ML is a field that has been weakly explored, but its potential has been described clearly. Using a controllable model for generating samples conditioned to some physical parameters can reduce biases beyond the imbalanced class issue.

\section{Method}
\label{Method}

From a high-level perspective, two main models interact in our approach: a variable star classifier and a deep generative model for generating synthetic light curves. Consequently,  Section \ref{classifier} provides a detailed exposition of the classifier model. Then, in Section \ref{secgenerative}, the PELS-VAE model is presented. Section \ref{sectraining} defines the interaction among all the involved models. After that, in Section \ref{secparameters}, the hyperparameter selection procedure is explained. Finally, Section \ref{implementation} covers the implementation aspects.

\subsection{Classifier model}
\label{classifier}

\begin{figure*}
\includegraphics[scale=0.6]{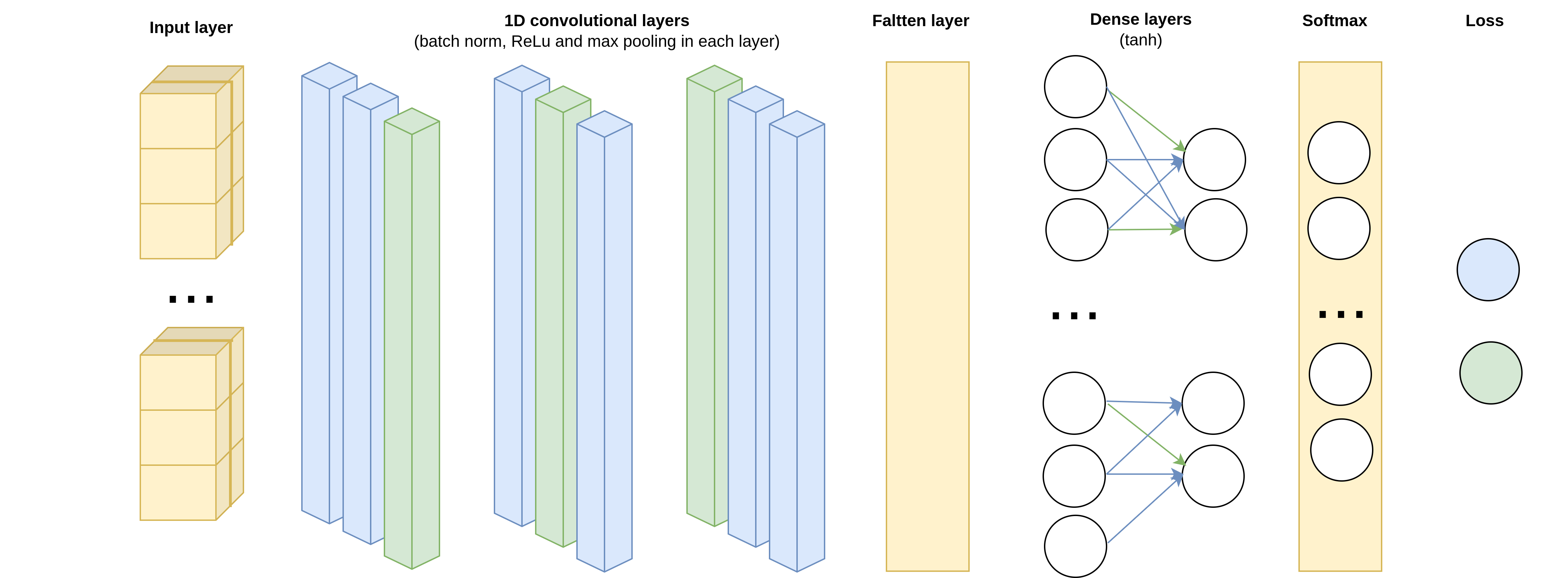}
\caption{Classifier model architecture. The input layer receives the $(\Delta t_i, \Delta m_i)_{i=1}^L$ representation. A sequence of 1D convolutional layers processes this input, where distinct sets of filters, managed by masks of weights, learn from real data ($\mathcal{L}_1$, depicted in light-blue) and synthetic data ($\mathcal{L}_2$, depicted in light-green), respectively. Each colour transformation corresponds to the filter operation it represents: light-blue for $\mathcal{L}_1$ and light-green for $\mathcal{L}_2$. }
\label{classifier-figure}
\end{figure*}

\color{black} In this work, we propose a classifier designed to process $(\Delta t_i, \Delta m_i)_{i=1}^L$  sequences, formatted as $2 \times L$ arrays, where $L$ is the sequence length. \color{black} The architecture, which is provided in Figure \ref{classifier-figure}, incorporates blocks of 1D convolutional layers, with the number of these blocks being a tunable hyperparameter from two to four. Each block comprises a 1D convolutional layer, followed by batch normalisation and a ReLU activation function. Additionally, a max pooling operator is integrated into each block to reduce the dimensionality and extract the most salient features. The first convolutional layer is equipped with 16 filters, each with a kernel size of six and a stride of one. Subsequent layers maintain the same kernel size and stride but double the filters in each new convolutional layer. After the convolutional layers, a fully connected layer is applied, the size of which depends on the number of convolutional layers due to the varying dimensions of the flattened output. The fully connected layer has an output dimension of 200. The network culminates with a final fully connected layer, which maps to the number of classes. We initialise the weights of all layers using the Xavier method \citep{glorot2010understanding}.

A well-known approach to regularise the training process is incorporating synthetic samples \citep{goodfellow2016deep}; however, some concerns must be considered to optimise the NN weights correctly. We highlight the balancing issue, which arises when there is a mismatch in the proportion of real and synthetic light curves used during training. An excessive number of synthetic light curves might lead the model to learn idealised data and inherit biases from the generation process, while too few may fail to provide sufficient regularisation. The main issue is that the number of synthetic light curves depends on the number of real light curves, which limits scalability. \color{black} For this reason, we propose training a dual CNN in which some filters are dedicated to learning real patterns from the training data, while others focus on capturing patterns from synthetic data. This design helps to avoid competition during the loss function minimization, alleviating the problem of finding the optimum synthetic-to-real data ratio. \color{black} In this way, the number of synthetic samples does not depend on the number of real light curves. This mask-based training approach emulates two systems: a primary system that learns from training data and a secondary system that learns from synthetic data with well-known patterns. The secondary system acts as a regularizer, constraining the model's parameters by learning from synthetic data. This method was recently employed to train more reliable feature-based multi-layer perceptron (MLP) classifiers, successfully mitigating the data shift problem in MLP models \citep{perez2023informative}.

Additionally, using a dedicated set of weights to learn synthetic light curves allows us to ensure the learning of the patterns injected synthetically. This approach helps to control the biases induced into the training set. \color{black}
We define weight masks to selectively update weights during learning, depending on the data type (real or synthetic). In Figure \ref{classifier-figure}, each mask is represented by a specific colour: light blue for \textit{mask}${1}$, which is designated for learning patterns from real light curves, and light green for \textit{mask}${2}$, which focuses on synthetic light curves. In convolutional layers, filters are fully assigned to one of the two masks based on a quantile threshold, whereas in fully connected layers, each weight is assigned independently. Specifically, filters/weights exceeding this threshold are allocated to \textit{mask}${1}$, which learns from the training set. In contrast, weights at or below this threshold are assigned to \textit{mask}${2}$, responsible for learning from synthetic data. This dual learning guides the solution space (weights) toward regions that are more likely to yield better generalisation, informed by the controlled integration of synthetic samples. This approach ensures that each set of weights is optimised for its corresponding data type. \color{black}
 
 To control the trade-off between learning from training data and synthetic light curves, we define two hyperparameters, $\varepsilon$ and $\nu$. The parameter $\varepsilon$ regulates the fraction of weights assigned to \textit{mask}$_{1}$, while $\nu$ scales the learning rate for \textit{mask}$_{2}$ relative to the learning rate of \textit{mask}$_{1}$. For example, if $\nu$ is set to 0.5 and the learning rate for \textit{mask}$_{1}$ is 0.1, then the learning rate for \textit{mask}$_{2}$ will be 0.05. Setting these parameters correctly, considering \textit{mask}$_{2}$ as a regularisation, is crucial for good training convergence.

Due to the size difference between the training and the synthetic sets, the weights associated with \textit{mask}$_{1}$ will updated more frequently than those in \textit{mask}$_{2}$. To increase the utilisation of synthetic light curves during learning, we allow the repetition of the forward and back-propagation procedures within a single epoch for synthetic batches. The number of iterations is controlled by a hyperparameter which is optimised (\texttt{repetitions}).  
\color{black}

\subsubsection{Loss functions}
Our architecture considers a different loss function for each mask of weights, i.e., for each type of light curve; therefore, in case of including parameters to control imbalanced class problems, each function can be customised. Three loss functions were proposed to assess the classifier and training robustness: cross-entropy, weighted cross-entropy and focal loss. 

\noindent \textit{Cross-entropy:} The cross-entropy loss function increases as the predicted probability diverges from the observed label. Thus, a perfect model would have a cross-entropy loss function of 0. The equation for a multi-class classification problem is given by
\[ \mathcal{L}_{\text{CE}} = - \sum_{i=1}^{C} y_i \log(p_i), \]
\noindent where \( C \) is the number of classes, \( y_i \) is a binary indicator (0 or 1) if class label \( i \) is the correct classification for the observation, and \( p_i \) is the predicted probability of the observation being of class \( i \).

\noindent \textit{Weighted cross-entropy:}  A variant of the standard cross-entropy loss is useful in managing imbalanced datasets. This loss function is adapted by incorporating a factor that scales the importance of each class. For example, if a class is rare, a higher weight is assigned, making the model more sensitive to errors in classifying this class; this loss function is expressed as
\[ \mathcal{L}_{\text{WCE}} = - \sum_{i=1}^{C} w_i y_i \log(p_i), \]
\noindent\noindent where, \( w_i \) represents the weight assigned to the \( i^{th} \) class. This weight is typically estimated as inversely proportional to the class frequency, giving more importance to less frequent classes. 
\color{black}

\noindent \textit{Focal loss:}  A loss designed to handle the issue of class imbalance by down-weighting well-classified examples. Its equation is given by
   \[ \mathcal{L}_{\text{Focal}} = - \sum_{i=1}^{C} \alpha_i (1 - p_i)^\gamma y_i \log(p_i), \]
\noindent where \( \alpha_i \) is a weighting factor for class \( i \), \( (1-p_i)^\gamma \) is a modulating factor with tunable focusing parameter \( \gamma \). The term \( (1 - p_i)^\gamma \) reduces the loss contribution from easy examples (where \( p_i \) is high) and increases the contribution from hard, misclassified examples.

\subsubsection{Performance metrics}

\noindent F$_1$:  This metric balances precision and recall by computing their harmonic mean. We utilise the macro-averaged F$_1$, which calculates F$_1$ for each class individually and then averages them, considering all classes equally. This is useful for assessing models in imbalanced datasets.

\noindent \textit{ROC-OVA:} The Area Under the Receiver Operating Characteristic Curve (AUC-ROC) evaluates the ability of a model to distinguish between classes. The one-vs-all approach computes the AUC-ROC for each class against all others, averaging by using the macro method, making it suitable for assessing multi-class classification. Its advantage lies in its simplicity and effectiveness in assessing each class independently, clearly understanding how well each class is differentiated. In this case, we calculate the macro average over five ROC curves, one for each star type.  

\noindent \textit{ROC-OVO:}  The one-vs-one AUC-ROC method takes a different approach. It compares every pair of classes, creating binary classifiers for each pair. The final AUC-ROC is the macro average of these comparisons, providing a detailed performance evaluation by considering the separability between all class pairs. By examining the separability of each class pair, this approach offers a comprehensive view of the model performance. ROC-OVO requires an average of $C(C-1)/2$ ROC curves, $C$ is the number of classes; it is to say, one for each possible pair of type of star, which amounts to ten curves in our case.

\subsection{Generative model}
\label{secgenerative}

The generative model proposed by \cite{martinez2022deep} is a conditional variational autoencoder (VAE) modified to inject knowledge about physical parameters using temporal convolutional blocks and reconstructing folded and normalised light curves. While the original PELS-VAE was trained with {\em Gaia} DR2, in this work, we train the model using data from {\em Gaia} DR3, which allows us to include additional physical parameters. This enhancement encourages us to incorporate six key physical parameters: period, absolute magnitude, effective temperature, radius, surface gravity, and metallicity.  

VAEs are generative models designed to model complex data distributions \citep{kingma2013auto}. Given a dataset \( X = \{x_1, x_2, \ldots, x_N\} \), a VAE defines a joint distribution over observed variables \( x \) and latent variables \( z \) as \( p(x, z) = p(x | z) p(z) \). The \color{black}loss function,\color{black} often referred to as the evidence lower bound (ELBO), is given by
\begin{equation}
\mathcal{L}(\theta, \phi; x) = \mathbb{E}_{q_\phi(z|x)}[\log p_\theta(x|z)] - D_{\rm  KL}(q_\phi(z|x) || p(z)),
\end{equation}
\noindent where \( \theta \) and \( \phi \) are the parameters for the decoder and encoder networks, respectively, $q_\phi(z|x)$ represents the approximate posterior distribution over the latent variables given the data and \( D_{\rm KL} \) is the Kullback-Leibler divergence. Traditional VAE has an important drawback known as the posterior collapse problem; this phenomenon implies that the VAE model ignores the latent variables and relies solely on the decoder \citep{lucas2019understanding}. A variant that allows managing this problem, improving, or disentangling the latent representation inducing a balance between reconstruction quality and regularisation is the $\beta$-VAE \citep{higgins2017beta, burgess2018understanding}; in $\beta$-VAE model, a new parameter $\beta$ controls the relevance of each component in the loss function as follows,
\begin{equation}
\mathcal{L}(\theta, \phi; x) = \mathbb{E}_{q_\phi(z|x)}[\log p_\theta(x|z)] - \beta D_{\rm KL}(q_\phi(z|x) || p(z)).
\end{equation}

\color{black} 
Despite the $\beta$-VAE model providing a more disentangled latent space, it is difficult to control the posterior distribution, for example, to explore how to change the posterior for different star types. To simplify this exploration, the conditional variational autoencoder (cVAE) was proposed, which extends VAEs by conditioning the model on additional information \( c \); typically, the labels \citep{richards2022conditional}. In cVAEs, the joint distribution is modelled as \( p(x, z | c) = p(x | z, c) p(z | c) \). The conditional ELBO becomes:

\begin{align}
\mathcal{L}\Big(\theta, \phi; x, c\Big) &= \mathbb{E}_{q_\phi(z|x, c)}\Big[\log p_\theta(x|z, c)\Big] \nonumber \\
&- \beta D_{\rm KL}\Big( q_\phi(z|x, c) \| p(z|c) \Big).
\end{align}

Furthermore, the PELS-VAE considers injecting physical parameters to enforce the relation between light curve features and stellar properties.  Temporal convolutional and dense layers are utilised in both the encoder and decoder. In the encoder, the physical parameters are incorporated after the temporal convolutional block and before the dense layers. \color{black} It is important to note that this model reconstructs normalised and folded light curves; hence, additional procedures, such as scaling and unfolding, are required to use the synthetic light curves created by PELS-VAE in a $(\Delta t, \Delta m)_{n = 1}^L$ input representation, as required by our classifier. \color{black} Several variants were tested to train this model, e.g. using imputation techniques to use the data from {\em Gaia} DR3 more efficiently, applying log transformation to the data sets, modifying $\beta$, etc. 
\color{black}

\subsection{Training process}
\label{sectraining}

\begin{figure*}
\includegraphics[scale=0.65]{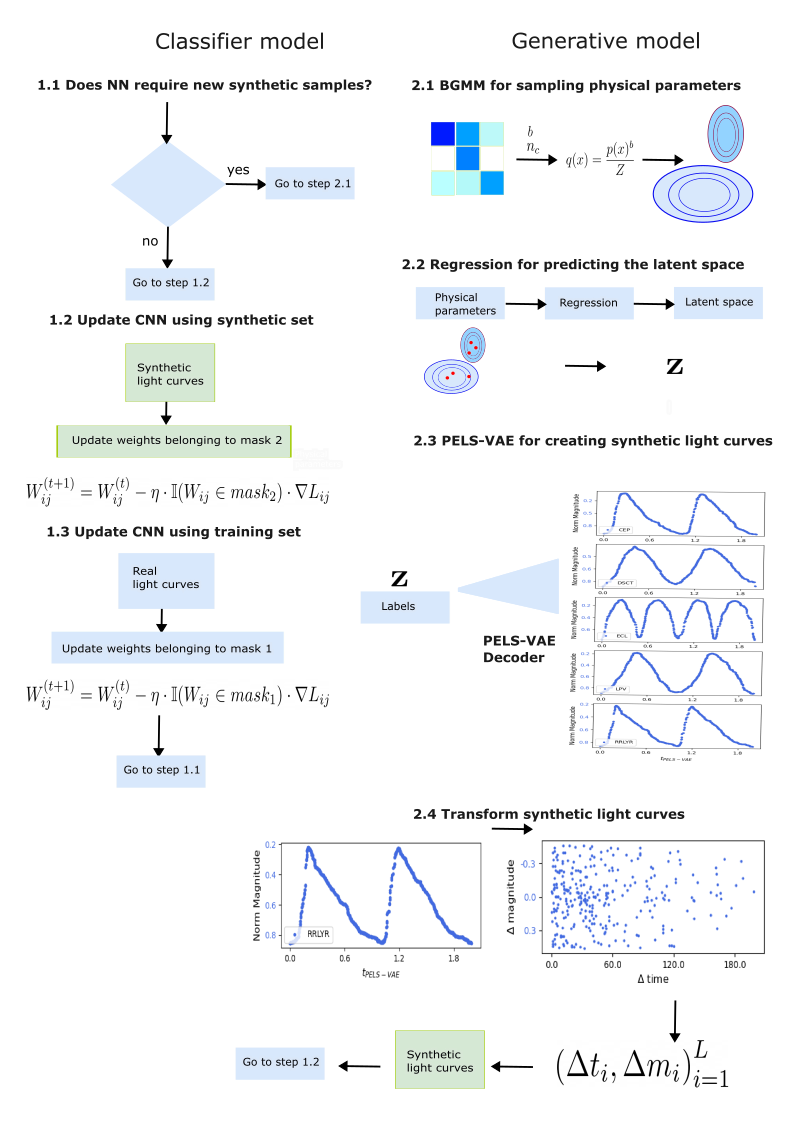}
\caption{Diagram illustrating one epoch of the proposed training overview. The process involves conditional checks for generating new synthetic samples, sampling strategies for obtaining synthetic physical parameters, multiple-output regression for predicting latent space, and transformations for obtaining representations used by the classifier model.}
\label{overviewmethod}
\end{figure*}

Figure \ref{overviewmethod} provides an overview of one training epoch of our proposed approach. A condition is checked in each epoch to decide if a new synthetic set of samples from the generative model is required (step 1.1). If a new sample is required, a sampling strategy is applied to obtain synthetic physical parameters from a Bayesian Gaussian mixture model (step 2.1). After that, using a multiple outputs regression, the latent space is predicted (step 2.2). Then, using this latent space, the physical parameters, and the label, we obtain a folded and normalised light curve via a trained PELS-VAE (step 2.3). Finally, the folded and normalised light curves are transformed to obtain the representation used for the classifier model (step 2.4).  In case no new synthetic samples are needed, first, an epoch is applied using the current synthetic light curves, updating only the weights belonging to \textit{mask}$_{2}$ (step 1.2). Afterwards, another epoch is applied using the training set with real light curves, with updates limited to weights belonging to \textit{mask}$_{1}$ (step 1.3). 

\subsubsection{Triggers of synthetic samples (step 1.1)}

At each epoch, the training procedure assesses whether there is a need to generate a new batch of synthetic samples. This decision is based on two conditions. The first condition, controlled by the hyperparameter \( E \), relates to the number of epochs elapsed without introducing new synthetic samples; it was set to three in the final experiments. The second condition, dictated by the hyperparameter \( \phi \), hinges on CNN's current proficiency in classifying synthetic samples. The hyperparameter \( E \) ensures a consistent introduction of diverse synthetic samples throughout the training process. Moreover, it is designed to gradually increase the generation of less probable light curves (reducing biases) as the training progresses. The hyperparameter \( \phi \), on the other hand, aims to increase the diversity and complexity of the synthetic data. It triggers the generation of new synthetic samples based on their classification difficulty, as indicated by the accuracy metric. Specifically, if the accuracy in classifying synthetic samples is excessively high, suggesting that they are too simple for the current model, it signals a need for more challenging samples. This hyperparameter was optimised in the hyperparameter search.

\color{black}
In addition, we need to determine how many objects will be sampled for each class. The policies presented below, which are based on the confusion matrix, were designed to define the number of samples from each variable star type. In the following section, we conduct experiments using each of these policies independently.
\color{black}

\noindent \texttt{Correct classification rate} (\texttt{CCR}): This policy calculates the \texttt{CCR} for each class, which is the ratio of correct predictions to the total number of instances for that class. \color{black} The classes are ranked based on these ratios in ascending order, considering a factor, \( r \), that reduces the number of objects according to the ranking. For example, if \( r = \frac{3}{2} \), the star class in position \( k+1 \) will receive \( \frac{2}{3} \) the number of objects compared to position \( k \). To calculate the number of objects per class, we define an additional hyperparameter, $\rm synthetic\_samples$. For instance, if we have five classes and set $\rm synthetic\_samples = 36$, the class in first place will receive 36 synthetic samples, while the second will receive 24. Moreover, we define a lower bound equal to \( \frac{1}{2} \times \rm synthetic\_samples \), ensuring that all classes receive a minimum allocation of synthetic samples. The next policies, \texttt{Max confusion} and \texttt{max\_pairwise\_confusion}
, use the same criteria to define the number of samples per class. \color{black}

\noindent \texttt{Max confusion}: This approach focuses on the classes most confused with others. Given a confusion matrix, it sums the off-diagonal elements in each row and ranks the classes based on these sums. 

\noindent \texttt{max\_pairwise\_confusion}: This procedure iteratively finds pairs of classes with maximum confusion and ranks the classes based on these pairs. 

\noindent \texttt{Proportion}: This method normalises the confusion matrix, considering off-diagonal elements, and then assigns a proportional budget of samples to each cell.

\noindent \texttt{non-priority}: Equal quantity of samples for each star type, \( \rm synthetic\_samples \).

\begin{figure*}
\includegraphics[scale=0.35]{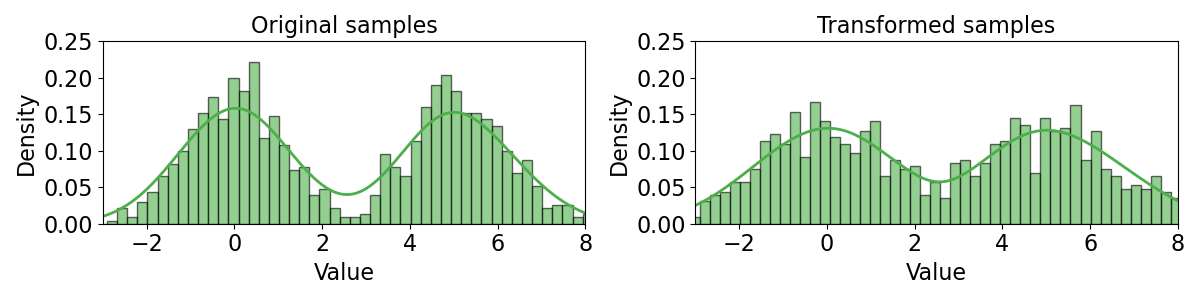}
\caption{Comparison of original and transformed samples from a GMM for a random variable that does not have physical meaning (Value). The histogram on the left shows the distribution of samples from the original distribution $(\mathcal{N} \left( \begin{bmatrix} 0 \\ 5 \end{bmatrix}, \begin{bmatrix} 1 & 0 \\ 0 & 1 \end{bmatrix} \right)$), and the histogram on the right shows the modified distribution using a $b = 0.6$, which accentuates the tails of the original distribution. Section \ref{gmmsection} dives into this distribution adaptation. }
\label{samplingbehaviour}
\end{figure*}

\subsubsection{Sampling method (step 2.1)}
\label{gmmsection}

The first step in the synthetic light curve generation involves sampling from the physical parameter space. \color{black} The number of objects per class, \( n_c \), is defined according to previously described policies. \color{black} To generate these \( n_c \) physical parameter samples, we use a Bayesian Gaussian mixture model (BGMM) for each variable star type. The reasons for selecting a BGMM include: (i) it allows us to inject expert knowledge about the means directly, preventing unstable behaviour when the class contains few examples; (ii) it eliminates the need to determine the optimal number of subpopulations; and (iii) each BGMM enables us to quantify zones that are more or less probable for physical parameters for each class. The means for each star type are also obtained from the expert knowledge (see Appendix \ref{physicalparameters}). \color{black} The BGMM, \( p(x) \), is an independent module, which is fitted prior to this learning process using physical parameters from the {\em Gaia} DR3 catalogue, as explained in detail in Section \ref{gaia}.  \color{black} For this fit, we use the variational inference approach \citep{blei2006variational}, implemented in \texttt{sklearn} \citep{pedregosa2011scikit}.

To control the biases, we define a modified probability density distribution \( q \) based on the target probability density distribution \( p \) as $ q(x) = \frac{p(x)^b}{Z}$. Here, \( Z \) represents a normalisation constant, calculated as \( Z = \int p(x)^b \, dx \). While direct evaluation of this integral might be computationally daunting, its exact value often becomes irrelevant in the sampling framework. The parameter \( b \) plays a pivotal role in the modified density. When \( b = 1 \), \( q \) mirrors \( p \), resulting in a traditional sampling process; if \( b > 1 \), the modified density accentuates differences, focusing sampling around \( p \)'s modes;
and finally, when \( 0 < b < 1 \), \( q \) explores  regions less represented in \( p \). Figure \ref{samplingbehaviour} shows a toy one-dimensional example of how sample distribution is modified by $b$, where $p(x)_{\rm 
 new}=\dfrac{p(x)^b}{Z}$. 

This sampling method is designed to generate synthetic light curves by varying physical parameters, enabling efficient exploration of the physical parameter space. The primary aim is to prioritise and represent less explored zones in the training dataset. The hyperparameter $b$ plays a crucial role in this process by adjusting the density distribution of these physical parameters. Specifically, when the value of $b$ is reduced, it effectively diminishes the peaks of the modes in the distribution. This modification ensures a more diverse and representative parameter space sampling, generating a wider variety of synthetic light curves.  An exponential decay function models the decay of $b$, $b = b_f + (1 - b_f) \exp(-c \times \text{epoch})$, where two additional hyperparameters, $b_f$ and $c$, must be defined. Intuitively, these hyperparameters model the minimum $b$ value by $b_f$, which sets the maximal modification of the initial probability density distribution and the speed of this decay process, controlled by $c$.

\subsubsection{Regression to latent space (step 2.2)}
\label{regression}

\color{black} The training of our regression model follows the pipeline outlined by \cite{martinez2022deep}. Specifically, we employ an RF model with multiple outputs, using the physical parameters extracted from {\em Gaia} DR3 as input features. The target variables of the regression are the latent space representations of stars, obtained by processing each star's data through the trained PELS-VAE encoder.

To optimise the hyperparameters—namely, the number of estimators, maximum depth, minimum number of samples in each split, and minimum number of samples in each leaf—a grid search with $k$-fold cross-validation is utilized. For performance evaluation, 20\% of the data is reserved for testing, and the mean absolute error is used as the error metric.

The primary purpose of this model is to predict the latent space based on a given set of physical parameters during each epoch when new synthetic light curves are generated. It is to say, after sampling the physical parameters, we use this regression model to predict the latent space before applying the PELS-VAE decoder. \color{black}

\subsubsection{Generate samples (Step 2.3)}

\color{black} Once the latent space is established, the decoder component of the PELS-VAE becomes essential for generating new synthetic light curves. A normalized light curve is generated by providing the PELS\_VAE decoder with a specific class and the corresponding latent space values, which are derived from physical parameters, as explained in Section \ref{regression}.
\color{black}

\subsubsection{ Adapt representation and create synthetic batch (Step 2.4)}
\label{adapt} 

The final stage of synthetic batch generation consists of three key steps: magnitude scaling, temporal sampling, and difference computation. During magnitude scaling, we select a real light curve from a star of the same type as those in the synthetic sample, choosing the one with the closest matching period. This real light curve is used to determine the maximum and minimum magnitudes, which are then applied to scale the synthetic light curve using the min-max scaling method as follows:
\[m^\prime = m_{\rm PELS-VAE} (m_{\rm max}-m_{\rm min}) + m_{\rm min} + \varepsilon,\]
\noindent where $m_{\rm max}$ and $m_{\rm min}$ are the maximum and minimum magnitudes from the real light curve, $m_{\rm PELS-VAE}$ is the normalized magnitude output by the PELS-VAE, and $\varepsilon \sim \mathcal{N}(\mu_p, \sigma^2_p)$ is a noise term, where $\mu_p$ and $\sigma^2_p$ represent the mean and variance of the photometric error.

Temporal sampling is performed by using the sampled period and calculating the number of cycles based on the baseline of the real light curve. The adjusted time, denoted as $t^\prime$, is computed as:
\[t^\prime = t_{\rm min} + k \times t_{\rm PELS-VAE},\]

\noindent where $k$ represents the number of cycles, $t_{\rm min}$ is the minimum time recorded in the real light curve and $t_{\rm PELS-VAE}$ is the temporal dimension which is generated by the PELS-VAE. For each observation, $k$ is sampled from a uniform distribution with bounds $a = 0$ and $b = $ the maximum number of cycles in the real light curve mentioned earlier.

Finally, we sort the observations in temporal order and compute the time and magnitude differences for each synthetic light curve, thus completing the synthetic batch generation process. Figure \ref{fig:panel} illustrates this process, from the PELS-VAE output to the CNN input.

\color{black} To train the classifier model, we use a sequence length of 300. \color{black} It is important to note that the PELS-VAE model is capable of generating sequences that are longer than those in the training set. Consequently, it is possible to sample multiple sequences from each synthetic light curve. The number of samples per synthetic light curve is governed by a hyperparameter, \texttt{n\_oversampling}. The choice of sampling method is particularly crucial for short sequences. If a random sampling approach is applied, there is a risk of omitting significant portions of the phase. In such cases, an equally spaced sampling strategy can be employed, with adjustments made to the first and last observations. \color{black}

\subsection{Hyperparameters}
\label{secparameters}

To manage this complex hyperparameter optimisation, we utilised a Bayesian optimisation framework implemented through \texttt{Weights \& Biases} to tune the hyperparameters \citep{wandb}. Two metrics were evaluated to identify the optimal parameters: the $F_1^{\text{val}}$ score and a weighted $F_1$ score, defined as: 
\[
\text{F}_1^{\rm weighted, \alpha} = (1-\alpha)*\text{F}_1^{\rm val} + \alpha*\text{F}_1^{\rm synthetic}.  
\]
$\text{F}_1^{\rm weighted, \alpha}$ score aims to leverage synthetic light curves to guide the exploration of the hyperparameter space, serving as a second layer in directing the learning process. In other words, we are applying regularisation to both the learning parameters (weights) and the hyperparameter optimisation. \color{black} We selected macro weighting for each F$_1$ score since it is more suitable for imbalanced classes; thus, during hyperparameter exploration, the search is directed toward models with better performance in scenarios with imbalanced classes. Given the extensive list of parameters and their range of alternatives, this hyperparameter search was supplemented with a grid-search exploration in preliminary stages, allowing for a better balance of exploration and intensification, i.e., exploration extensively searches the solution space to avoid local optima while intensification exhaustively searches promising regions to find the best solution \citep{yang2014metaheuristic}.

\subsection{Implementation}
\label{implementation}
Our algorithms were implemented using \texttt{Pytorch}. Bayesian optimisation was applied from the \texttt{Weights \& Biases}\footnote{\href{https://wandb.ai/}{https://wandb.ai/}} platform. The PELS-VAE was trained using the original source code. The complete source code can be found in \href{https://github.com/frperezgalarce/cnn-pels-vae}{https://github.com/frperezgalarce/cnn-pels-vae}. Note that in this code, synthetic batches can be generated independently, which can be useful for injecting synthetic samples during training in new classifiers using both pre-trained PELS-VAE and pre-trained BGMM.  Data sets, hyperparameter configurations used to train this model, and trained models are available in \href{https://zenodo.org/records/13284675}{https://zenodo.org/records/13284675}.

\section{Data}
\label{Data}

\subsection{OGLE}

The classification approach was tested using the OGLE-III online catalogue of variable stars\footnote{\href{ https://ogledb.astrouw.edu.pl/~ogle/CVS/}{https://ogledb.astrouw.edu.pl/~ogle/CVS/}}, which is part of the Optical Gravitational Lensing Experiment \citep[OGLE;][]{udalski2008optical}. The OGLE project is a long-term project aimed at detecting microlensing events and identifying exoplanets via transit methods, including observations from critical astronomical areas such as the Galactic bulge, the Large and Small Magellanic Clouds, and the Carina constellation. To train and test our model, we utilised 419,257 light curves from the I band, conducting various experiments to assess the impact of different, training set size, signal-to-noise ratio levels and sequence lengths.

\subsection{{\em Gaia} DR3}
\label{gaia}
\begin{figure*}
\includegraphics[scale=0.35]{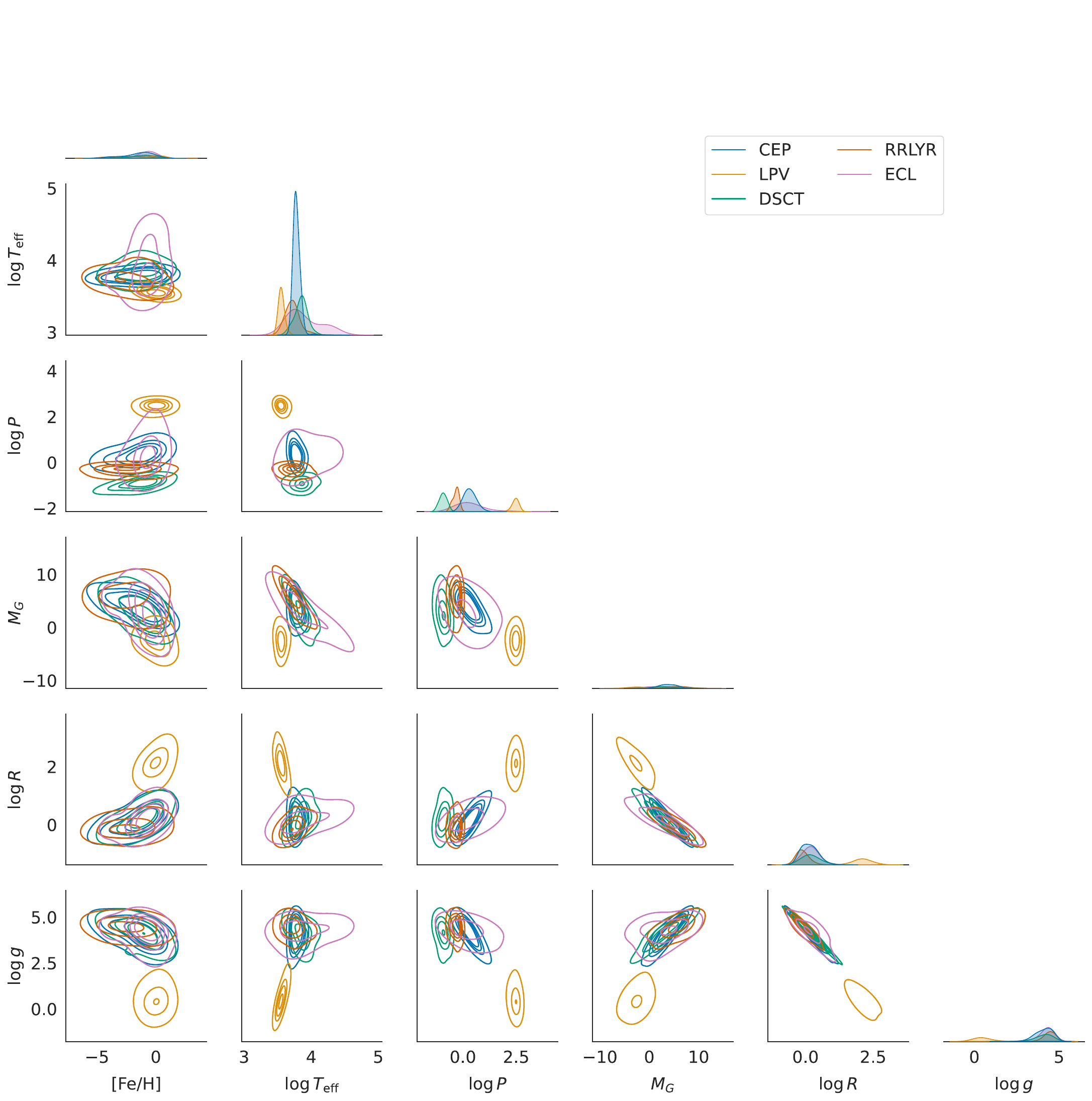}
\caption{\color{black} Physical parameters by class extracted from {\em Gaia} DR3. \color{black} Each color represents a star type, as indicated in the legend. CEP represents Cepheids, DSCT represents delta Scuti stars, ECL represents eclipsing binaries, LPV represents long period variables, and RRLYR represents RR Lyrae stars.  \color{black} }
\label{physical_parameters}
\end{figure*}

In the training of our PELS-VAE model, we utilise six key physical parameters derived from the {\em Gaia} DR3 catalogue, one of the most comprehensive resources for astrophysical data available \citep{creevey2023gaia}. The parameters considered are metallicity ([Fe/H], measured in dex), period ($P$, in days), absolute magnitude in the G band ($M_G$), surface gravity ($\log g$, in dex), radius ($R$, in $R_\odot$), and effective temperature ($T_{\rm eff}$, in K).  While we specifically use the ${\rm [Fe/H]_{\rm J95}}$ metallicity scale \citep{jurcsik1995revision}, we refer to it simply as [Fe/H] in subsequent sections. We applied logarithmic transformations to the effective temperature, period, and radius to scale the data distribution. Figure \ref{physical_parameters} shows the distribution of these parameters.

To integrate observational data from {\em Gaia} DR3 with OGLE, we utilised the cross-matching technique described by \citet{martinez2022deep}, successfully matching 53,090 stars. We decided to impute physical parameters after preliminary experiments during the PELS-VAE training with and without imputation. Imputation was performed stratified by class using the $k$-nearest neighbour (KNN) method \citep{troyanskaya2001missing} with $k=5$. This method not only allows us to use a bigger training set but also ensures that imputed values represent the nearest neighbours within the same class, avoiding confusion with objects from different classes and similar physical parameters.  Table \ref{missingdata} details the missing data within the physical parameters. 

When examining Figure \ref{physical_parameters}, which shows the physical parameters used to enhance the PELS-VAE, it becomes clear that the stellar classes exhibit distinct characteristics in these physical parameters that encourage us to start the synthetic curve generation in this space. The univariate period distribution is particularly informative; four stellar classes can be separable. Eclipsing binaries tend to overlap with other categories, which could indicate similarities in their physical parameters. Specifically, the delta Scuti stars (DSCT) are noticeable in the lower period range. Following these, the RR Lyrae stars are observable, with the Cepheid stars and then the long period variables being identifiable at progressively higher periods. In the bivariate log radius versus log period plot, a well-known pattern appears where the Cepheid stars display a pronounced positive correlation, showcasing a direct proportionality between their pulsation periods and radius. This relationship is consistent with the established period-luminosity relation. When we delve into the multivariate space – considering other parameters such as metallicity, effective temperature, absolute G magnitude, and surface gravity – the separability of these classes does not always become more pronounced. Long-period variable (LPV) stars are separable for all these physical parameters in this set. The data depicted in Figure \ref{physical_parameters} can be complemented by the traditional stellar classifications of the Hertzsprung-Russell diagram by showing variable stars with diverse periods, radii, and luminosities. It confirms the established correlations, such as the period-luminosity relationship for Cepheids.

\subsection{Induced biases}
\label{inducedbiases}

Two data shift scenarios were designed to validate our proposal. These benchmark sets aim to replicate the inherent biases present in the training data of variable stars. To generate these benchmark sets, we employ an active learning metric based on uncertainty sampling; specifically, we utilise the Gini impurity index ($G_{\rm index}$), which is defined as $G_{\rm index} = 1 - \sum_{k=1}^{|K|} p_k^2$,  where \( |K| \) represents the number of classes and \( p_k \) is the probability of selecting an element from class \( k \). For a given instance, the $G_{\rm index}$ quantifies how mixed the class probabilities are. If an instance has a high $G_{\rm index}$, the model is more uncertain about which class the instance belongs to.
 
To obtain \( p_k \), we first have to fit a probabilistic classifier; it has to be trained on the entire catalogue to generate soft predictions for each star, representing the probability of belonging to a particular class. Using these probabilistic outcomes, we can quantify the uncertainty level for each star, which can be interpreted in our context as the classification difficulty. To facilitate the allocation of objects to the training or testing sets without employing a strict threshold (i.e., $\text{if } G_{\rm index} \leq \text{threshold} \rightarrow \text{train} \text{ else testing}$), we incorporate randomness into the selection process, modulated by a tempering constant \( T \) as follows. For each star, we first compute the probability $p = e^{-G_{\rm index}/T}$ and then sample a random value $r$ from a uniform distribution between 0 and 1. If $r \leq p$, the data point is added to the training dataset; otherwise, it is added to the testing dataset. In this way, uncertain objects are more likely to be included in the testing set. The parameter \( T \) controls the degree of shift in the dataset, with higher \( T \) values yielding more balanced sets.

The first induced bias, which generates \textit{Data set I}, involves training an RF RR Lyrae binary classifier using features estimated through the FATS package \citep{nun2015fats}. In this case, the bias is introduced by focusing on including stars that are more likely to be uncertain and could be confused with RR Lyrae stars in the testing set; for this scenario, we used \( T = 1 \). The second case, which generates \textit{Data set II}, considers an RF multi-class classifier. In this scenario, we used \( T = 3 \). The bias in this scenario is designed to induce confusion among all classes by introducing ambiguity in class labels across the dataset. This approach helps assess how well the model can generalise when faced with data shifts involving multi-class overlapping in the testing set. Table \ref{tab:trainingset_combined} presents the distribution of objects within the testing and training sets for each induced bias.

We used the Mann-Whitney U test \citep{macfarland2016mann}, a univariate non-parametric method, to statistically validate the differences in period and amplitude between the training and testing sets for both datasets and across all classes. Specifically, the null hypothesis ($H_0$) states that the two populations are identical. In the majority of cases, the results rejected the null hypothesis, indicating significant group differences; the only exception is for DSCT stars in the period feature, where the null hypothesis was not rejected. This result is likely influenced by the limited number of samples available for this star type. Given these results, we confirm that the induced biases within the benchmark data sets contain the data shift problem, making them suitable for testing our approach.

\begin{table*}
\centering
\caption{Number of objects and percentage per class in training and testing sets for Data set I and Data set II. The validation set is obtained from the training set (30\%) in a stratified sample.}
\label{tab:trainingset_combined}
\begin{tabular}{lllcccc}
\hline
& Class  & Class name & \multicolumn{2}{c}{\textit{Data set I}} & \multicolumn{2}{c}{\textit{Data set II}} \\
&        &            & Training (occurrences, \%) & Testing  (occurrences, \%) & Training  (occurrences, \%) & Testing  (occurrences, \%) \\ \hline
\textit{} & CEP   & Cepheids                & 4,093 (1.05\%)  & 3,859 (13.63\%) & 4,472 (1.14\%)  & 3,480 (12.23\%) \\
          & DSCT  & Delta Scuti             & 1,496 (0.38\%)  & 1,311 (4.63\%)  & 1,447 (0.37\%)  & 1,360 (4.78\%)  \\
          & ECL   & Eclipsing Binaries      & 35,061 (8.97\%) & 6,726 (23.76\%) & 34,652 (8.87\%) & 7,135 (25.08\%) \\
          & LPV   & LPVs                    & 321,451 (82.22\%) & 2,527 (8.93\%)  & 316,092 (80.88\%) & 7,886 (27.72\%) \\
          & RRLYR & RR Lyrae                & 28,847 (7.38\%)  & 13,886 (49.05\%) & 34,144 (8.74\%)  & 8,589 (30.19\%) \\
\hline
& Total  &                           & 390,948         & 28,309          & 390,807         & 28,450           \\

\hline
\end{tabular}
\end{table*}

\section{Results}
\label{Results}

\subsection{Hyperparameter selection}

Given that the data shift problem is crucial in our experimental setting, it is unclear if traditional performance metrics like F$_1$ or accuracy on the validation set work well in exploring the hyperparameter space. Because of that, we compare a traditional approach using an F$_1^{\rm val}$ and $\text{F}_1^{\rm weighted, \alpha}$. In the $\text{F}_1^{\rm weighted, \alpha}$, synthetic light curves were considered during the hyperparameter exploration but were not used to calculate the score for selecting the best hyperparameter set. 

\color{black} Table \ref{table:f1_scores} compares the three search methods across both benchmark sets, including the best-found values during each exploration (BF), the recommended hyperparameters (RH), and the policy methods that achieved these values. \color{black} The objective score that provided the best results in both sets was the weighted F$_1^{0.15}$, indicating that incorporating synthetic samples to define model performance can help to explore the hyperparameter space. The hyperparameter search yielded better results not only in the RH but also in the BF.  This empirical evidence encourages paying more attention to hyperparameter optimisation in light of underlying data issues in variable stars, such as data shift or class imbalance problems.

\begin{table*}
\centering
\caption{Hyperparameter optimisation. RH: objective score for recommended hyperparameters, BF: best found objective score, BFP: policy used in best-found solution, RFP: the policy used in the recommended solution.}
\begin{tabular}{llccll}
\hline
Data set & Exploration & BF & RH & BFP & RFP \\ \hline
\textit{Data set I} & F$_1{\rm val}$ & 0.56 & 0.52 & \texttt{non-priority}  & \texttt{max\_pairwise\_confusion} \\ 
& F$_1^{\rm weighted, 0.15}$ & \textbf{0.58}& \textbf{0.54} & \texttt{Max confusion} & \texttt{max\_pairwise\_confusion} \\ 
& F$_1^{\rm weighted, 0.3}$ & 0.57 & 0.52 & \texttt{Proportion} & \texttt{max\_pairwise\_confusion} \\ \hline
\textit{Data set II} & F$_1^{\rm val}$ & 0.65 & 0.65 & \texttt{Proportion}  & \texttt{Proportion} \\ 
& F$_1^{\rm weighted, 0.15}$ & \textbf{0.67} & \textbf{0.66} & \texttt{Max confusion} & \texttt{max\_pairwise\_confusion} \\ 
& F$_1^{\rm weighted, 0.3}$ & 0.66 & 0.65 & \texttt{Proportion} & \texttt{CCR} \\ \hline
\end{tabular}
\label{table:f1_scores}
\end{table*}

\subsection{Synthetic light curves}
\begin{landscape}
\begin{figure}
        \begin{minipage}{0.24\linewidth}
        \includegraphics[width=\linewidth]{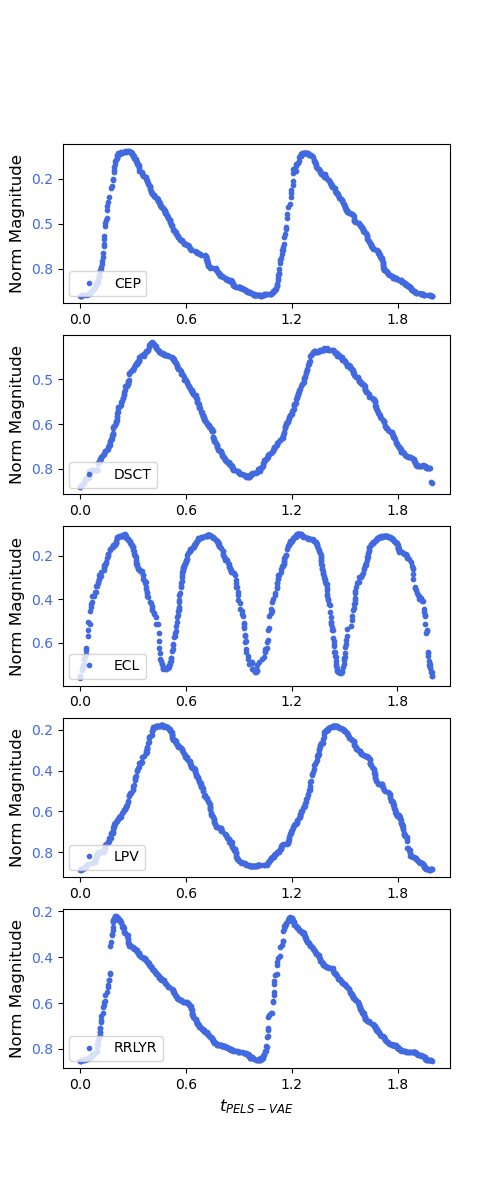}
    \end{minipage}
    \begin{minipage}{0.24\linewidth}
        \includegraphics[width=\linewidth]{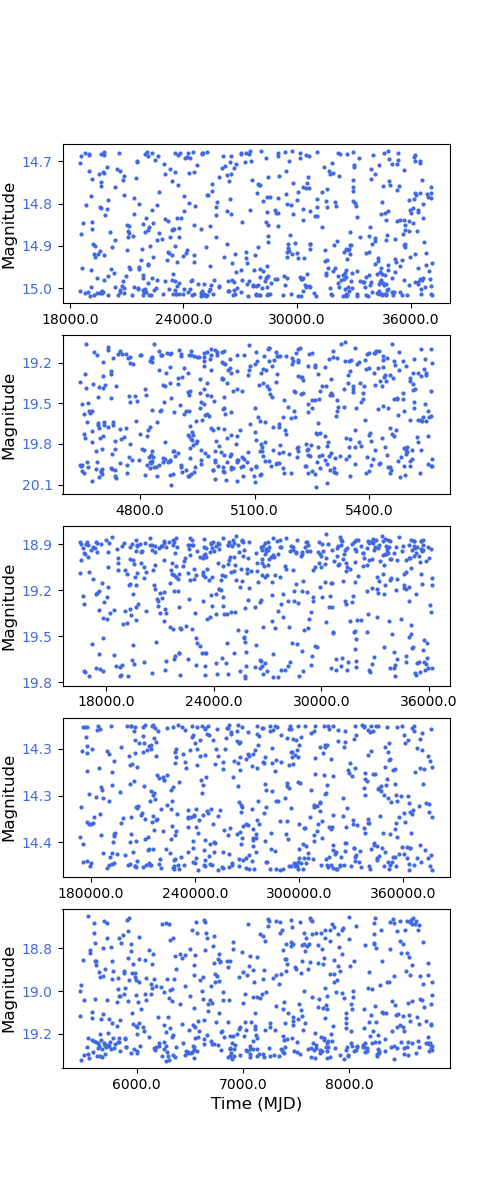}
    \end{minipage}    
    \begin{minipage}{0.24\linewidth}
        \includegraphics[width=\linewidth]{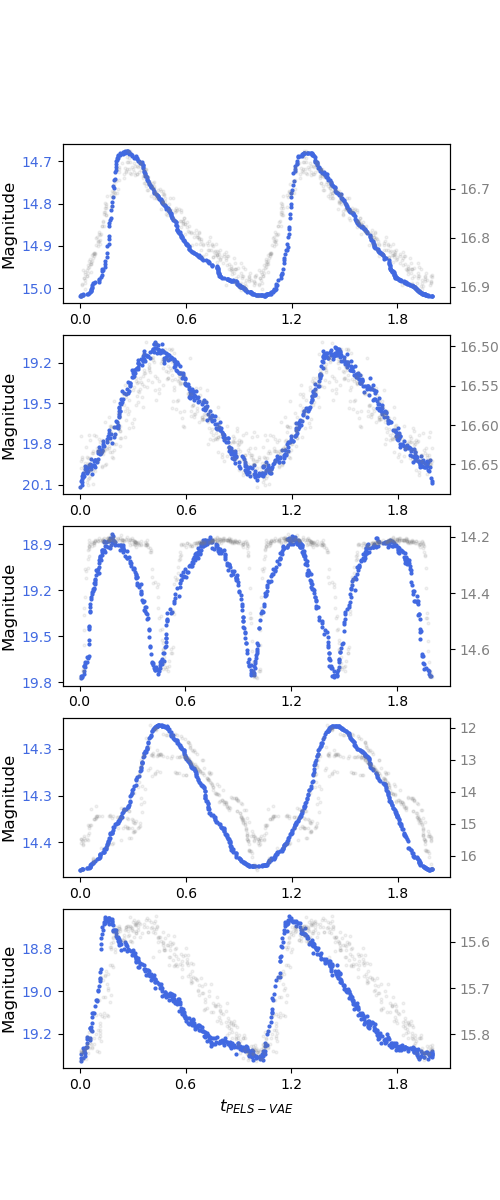}
    \end{minipage}
    \begin{minipage}{0.24\linewidth}
        \includegraphics[width=\linewidth]{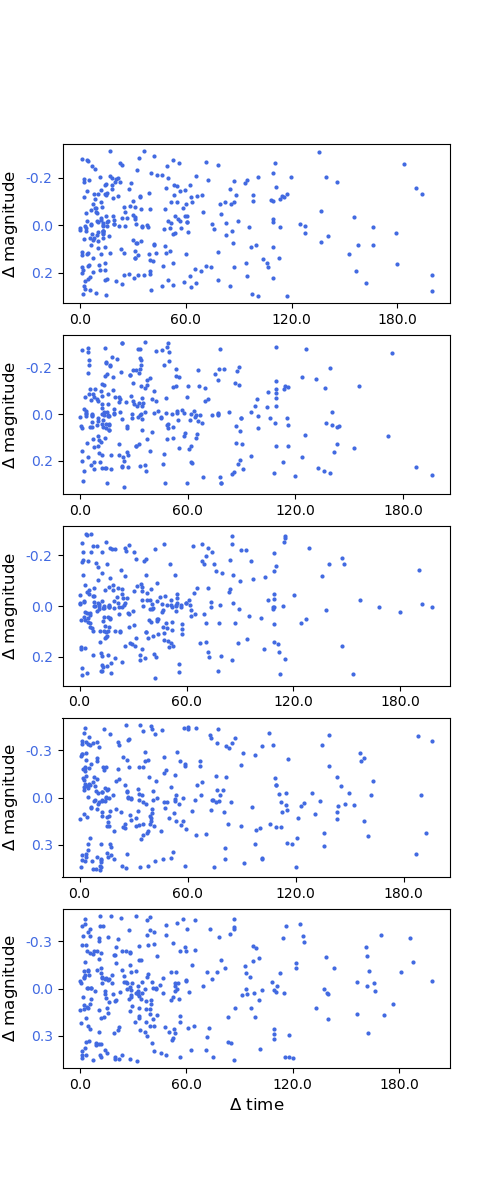}
    \end{minipage}
    \caption{Dataflow for synthetic light curves. The first column shows the light curves predicted by the PELS-VAE for these samples, i.e., the normalised and phased light curve outputted from the PELS-VAE. The second column presents a reverted light curve according to the procedure explained in Section \ref{adapt}. The third column illustrates a synthetic folded light curve (blue) with respect to a light curve in the training set (grey) with similar physical parameters to validate the process and visualise the impact of added noise. Finally, the last column provides the $(\Delta t, \Delta m)_{n=1}^L$ representation. Each row corresponds to a different variable star type (Cepheids, delta Scuti, eclipsing binaries, LPV and RR Lyrae). }
    \label{fig:panel}
\end{figure}
\end{landscape}

\begin{landscape}
\begin{figure}
        \begin{minipage}{0.24\linewidth}
        \includegraphics[width=\linewidth]{plot_magnitudet_pels-vae_ID_167.png}
    \end{minipage}
    \begin{minipage}{0.24\linewidth}
        \includegraphics[width=\linewidth]{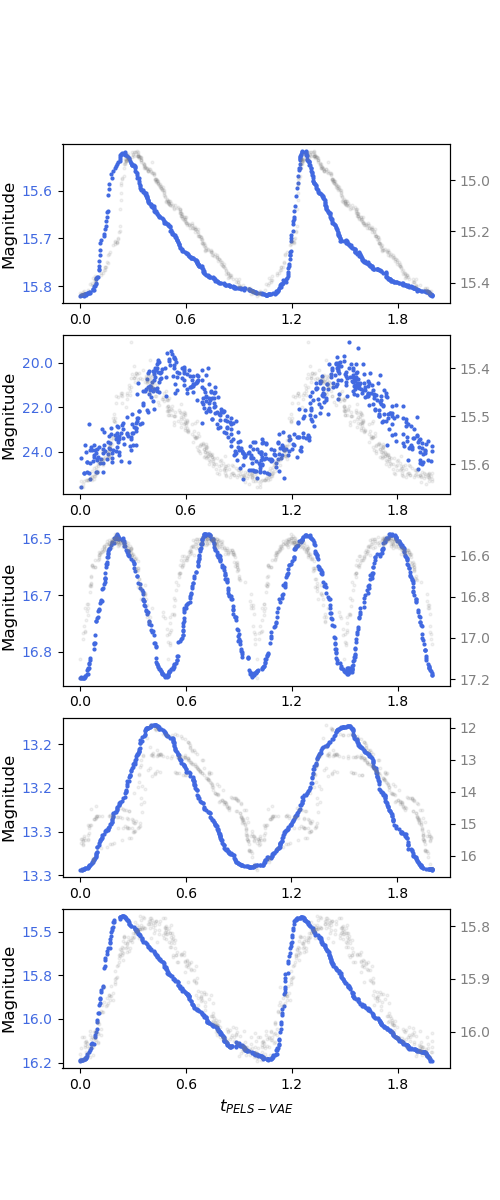}
    \end{minipage}    
    \begin{minipage}{0.24\linewidth}
        \includegraphics[width=\linewidth]{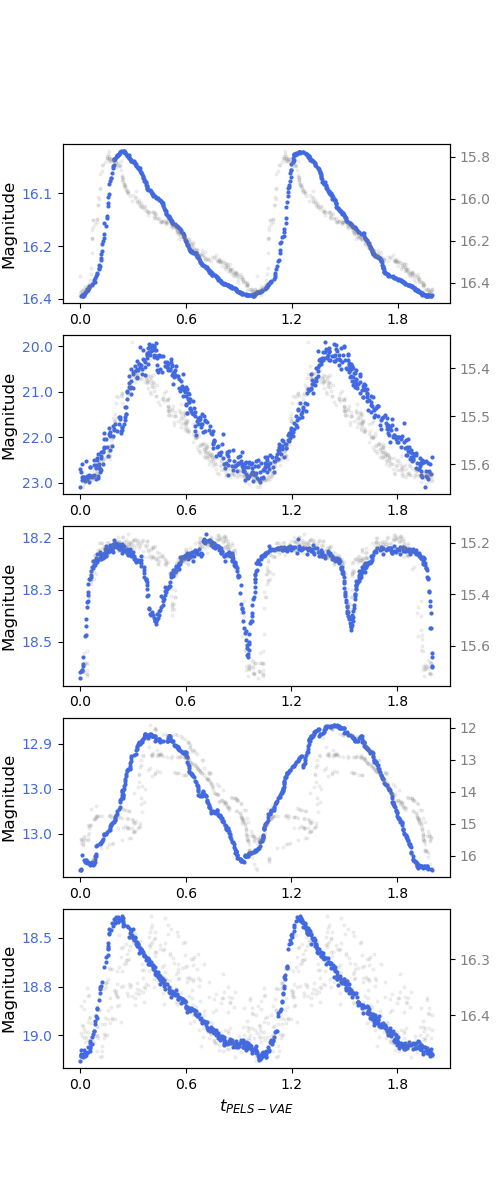}
    \end{minipage}
    \begin{minipage}{0.24\linewidth}
        \includegraphics[width=\linewidth]{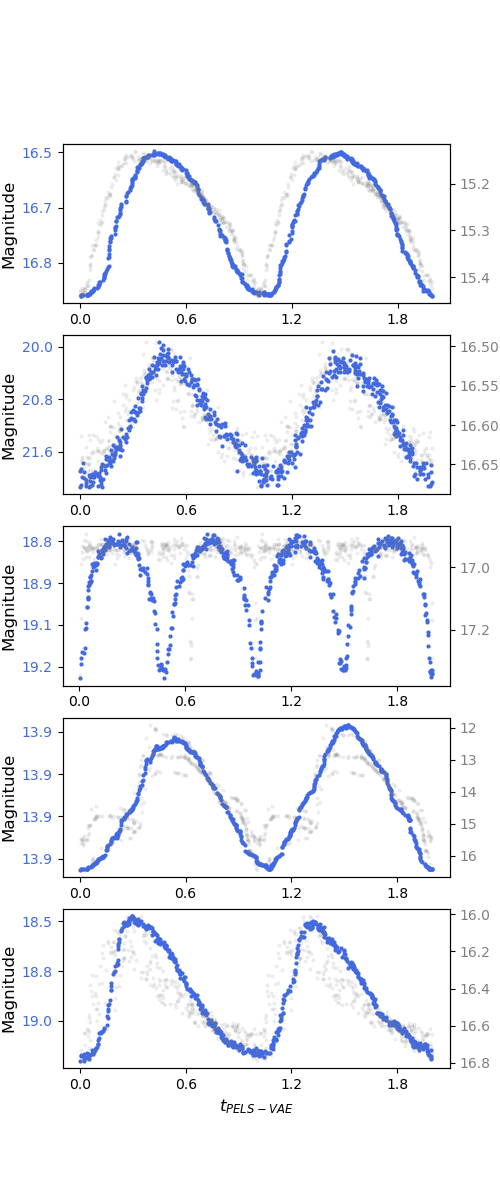}
    \end{minipage}
    \caption{Different sample of synthetic light curves compared with the closest real light curves, matched using k-nearest neighbors (kNN) based on physical parameters. The first column shows the light curves predicted by the PELS-VAE for these samples, i.e., the normalised and phased light curve outputted from the PELS-VAE. The second column presents a reverted light curve according to the procedure explained in Section \ref{adapt}. The third column illustrates a synthetic folded light curve (blue) with respect to a light curve in the training set (grey) with similar physical parameters to validate the process and visualise the impact of added noise. Finally, the last column provides the $(\Delta t, \Delta m)_{n=1}^L$ representation. The star categories are ordered from top to bottom as follows: Cepheids, delta Scuti, eclipsing binaries, long period variables, and RR Lyrae stars. }
    \label{fig:synthetic_light_curves2}
\end{figure}
\end{landscape}

This section focuses on providing an example of the data pipeline from physical parameter samples to the input representation fed to the classifier. It is worth highlighting that this data pipeline can be used independently for another classifier as a tool for synthetic data generation; even using a different input representation, it only receives the $b$ parameter as input, which controls the mode peaks when sampling physical parameters. Figure \ref{fig:panel} shows the dataflow from light curves predicted by the PELS-VAE to the $(\Delta t, \Delta m)_{n=1}^L$ representation. The first column displays the synthetic light curves generated by the PELS-VAE. The second column presents the same light curves, converted into raw light curves as explained in Section \ref{adapt}. The third column shows the folded light curves, including photometric errors, alongside a real light curve from the training set with similar physical parameters, identified using KNN; this column is displayed solely to validate the conversions. Finally, the last column shows the  $(\Delta t_i, \Delta m_i)_{i=1}^L$ representation, which is the input for our proposed classifier. Additional examples of synthetic light curves, compared with the closet object from the training set, are presented in Figure \ref{fig:synthetic_light_curves2}. It is important to note that replicating real light curves is not our objective. \color{black} We compare the synthetic light curves with similar real light curves in the physical parameter space. To be clear, the exact set of physical parameters used for the generated light curves does not exist within our training set; instead, they are sampled from the physical parameter distribution (see Section \ref{gmmsection}). We highlight that replicating an existing light curve from the training set is less challenging than generating light curves from unseen physical parameters, as the model has already encountered this light curve and can learn it nearly perfectly. \color{black} However, such light curves are not useful for addressing the data shift problem. Our primary focus is on generating synthetic light curves from underrepresented regions of the physical parameter space, where data is scarce. Although the synthetic light curves are not perfect, the patterns for each star type are clear, which has enabled us to improve the classifier's reliability.

\subsection{Comparison of policies and loss functions}

Table \ref{policiesresults} presents the experimental results for different policies, including the mean, minimum, and maximum values for F$_1$, ROC-OVO, and ROC-OVA. For \textit{Data set I}, the proportional policy outperforms the other policies across all metrics. In the case of \textit{Data set II}, the \texttt{no\_priority} policy achieves the highest value in seven of the nine metrics. The maximum values for the three metrics show higher variability, as the maximum value for F$_1$ was obtained by the \texttt{max\_pairwise\_confusion} policy, the maximum value for ROC-OVO was achieved by the \texttt{CCR} (and \texttt{no\_priority}) policies, and the maximum value for ROC-OVA was found by the max\_confusion policy. When examining Table \ref{policiesresults}, significant performance differences can be observed when different policies are used to assign the number of synthetic stars, underscoring the importance of policy selection. Moreover, the varying results across the two different data sets emphasise the need to choose policies based on the specific characteristics of each data set. The \texttt{proportion} policy and \texttt{no\_priority} policy will be used in the following sections for experiments with \textit{Data set I} and \textit{Data set II}, respectively. 

Table \ref{performancelossfunctions} shows results for different loss functions using our training approach on the \textit{Data set I}. Despite some differences in these metrics, all loss functions perform well within this training approach. Cross entropy, which does not consider weights for managing the imbalance problem, exhibits higher variability in results, obtaining several worst performances, but it also achieves the best performance in other metrics. Focal loss and weighted cross entropy never obtain the worst performance and achieve some of the best performances. Weighted cross entropy achieved the best results in all metrics related to the F$_1$, while focal loss obtained the best results in both minimum values for the ROC-based metrics. In the following experiments, we will use the focal loss function due to its greater flexibility in the hyperparameter search and performance comparable to that of weighted cross-entropy.

\subsection{Signal-to-noise ratio and sequence length impact}

\color{black} When examining Table \ref{sn_ration_an}, we observe that incorporating synthetic samples and mask-based training, referred to as "Two losses" or self-regulated training, leads to improvements in nearly all performance metrics compared to training without regularization masks (referred to as "One loss"). For \textit{Data set I}, the F$_1$ score increases or remains equal in all experiments, with a maximum average increase of 1.5. In general, improvements are observed in 3 out of 4 cases, one of which is statistically significant. Note that $^{\ast}$ indicates that the samples come from different distributions according to the Mann–Whitney U rank test, while $^{+}$ denotes a statistically significant difference according to the Student’s t-test. Regarding ROC metrics, the mean increase in ROC-OVO ranges from 0.2 to 0.9, while the mean increase in ROC-OVA ranges from 0.6 to 0.9.  Similarly, in Table \ref{sn_ration_an}, we observe that applying our training approach with synthetic samples leads to consistent improvements for \textit{Data set II}. The F$_1$ score increases in three out of four cases. In the remaining case, there is a slight decrease, but it is not statistically significant. In contrast, two of the improvements are statistically significant, with mean F$_1$ differences of 0.6 for sn\_ratio equal to 6 and 0.9 for sn\_ratio equal to 10. The ROC metrics for \textit{Data set II} also show a clear trend of improvement. The self-regulated training performs equal to or better than the standard training in all cases. The average metric improvements are statistically significant in almost every scenario. Table \ref{sn_ration_an} also shows that classification performance remains stable even as the signal-to-noise ratio decreases, indicating that our approach effectively handles high noise levels in observations without compromising performance metrics. 

Table \ref{seqlength} reports the mean, minimum, and maximum values for F$_1$, ROC-OVO, and ROC-OVA across two sequence lengths (50 and 150). Our regularised approach consistently performs competitively across most performance metrics. For \textit{Data set I}, incorporating two loss functions led to a statistically significant improvement of 2.9 points in the F$_1$ score when using 50 observations. In addition, the minimum F$_1$ score increased from 38.1 to 43.8. Similarly, both the ROC-OVO and ROC-OVA metrics showed improvements under the two-loss configuration with 50 observations. However, when 150 observations are used, no statistically significant differences are observed. For \textit{Data set II}, we also do not observe any significant differences.

\color{black}

\begin{table*}
\centering
\caption{Statistics for policies and sample size in testing sets with mean, minimum, and maximum values. The highest values in each column are highlighted in bold. \color{black} A repeated holdout with ten repetitions was used to obtain these metrics, and 40,000 light curves were included in the training set for each experiment.\color{black} }
\begin{tabular}{lrlrrrrrrrrr}
\toprule
{} &  &          Policy & \multicolumn{3}{c}{F$_1$} & \multicolumn{3}{c}{ROC-OVO} & \multicolumn{3}{c}{ROC-OVA} \\
{} &     &  &  mean & min &   max &         mean &   min &   max &         mean &   min &   max  \\
\midrule
 \textit{Data set I} &       &                     \texttt{CCR} &    56.4 &  53.7 &  59.3 &         88.0 &  85.3 &  89.6 &         87.0 &  85.4 &  88.7 \\
 &        &           \texttt{max\_confusion} &    56.5 &  53.4 &  59.5 &         88.2 &  85.9 &  89.2 &         87.1 &  84.6 &  88.1 \\
 &        &  \texttt{max\_pairwise\_confusion} &    56.7 &  52.2 &  60.9 &         88.2 &  86.5 &  90.1 &         87.1 &  85.2 &  89.2 \\
 &        &             \texttt{no\_priority} &    57.3 &  54.4 &  60.1 &         88.6 &  87.5 &  89.9 &         87.5 &  86.1 &  88.6 \\
 &        &              \texttt{Proportion} &    \textbf{57.8} &  \textbf{54.6} &  \textbf{61.9} &         \textbf{88.8} &  \textbf{87.7} &  \textbf{90.6} &         \textbf{87.8} &  \textbf{86.9} &  \textbf{89.6} \\
\hline
\color{black}  \textit{Data set II}  &   & \texttt{CCR} &    57.5 &  52.2 &  61.4 &         88.9 &  85.1 &  \textbf{90.4} &         89.9 &  86.6 &  91.2 \\
 &        &           \texttt{max\_confusion} &    58.1 &  52.7 &  61.2 &         88.8 &  86.3 &  91.0 &         89.8 &  87.4 &  \textbf{91.7} \\
 &        &  \texttt{max\_pairwise\_confusion} &    58.1 &  51.2 &  \textbf{62.9} &         88.8 &  86.4 &  90.3 &         89.7 &  87.6 &  91.0 \\
 &        &             \texttt{no\_priority} &  \textbf{59.0} &  \textbf{54.5} &  62.3 &         \textbf{89.2} &  \textbf{88.0} &  \textbf{90.4} &         \textbf{90.0} &  \textbf{89.1} &  91.1 \\
 &        &               \texttt{Proportion} &    57.7 &  49.3 &  62.0 &         88.3 &  82.9 &  90.1 &         89.3 &  85.0 &  90.8 \\
\bottomrule
\end{tabular}
\label{policiesresults}
\end{table*}

\begin{table*}
\caption{Classification performance metrics in testing sets for different loss functions across ten experiments. 40,000 light curves were considered in the training set for each experiment. Red and black numbers indicate the worst and best performance for each metric, respectively.}
\begin{tabular}{llrrrrrrrrr}
\toprule
{} &                           Loss function & \multicolumn{3}{c}{F$_1$} & \multicolumn{3}{c}{ROC-OVO} & \multicolumn{3}{c}{ROC-OVA} \\
{} &  &   mean &   min &   max &         mean &   min &   max &         mean &   min &   max\\
\midrule
 &  Weighted cross entropy& \textbf{57.6} & \textbf{55.4} & \textbf{62.4} & \textbf{\color{red}88.5\color{black}} &  86.8 &  89.7 & 87.5 & 85.7 &  89.0 \\
 & Focal  & 56.7 & 52.4 &  \textbf{\color{red}60.0\color{black}} & \textbf{\color{red}88.5\color{black}} &  \textbf{86.9} &  \textbf{\color{red}89.6\color{black}} &   \textbf{\color{red}87.4\color{black}} &  \textbf{85.9} &  \textbf{\color{red}88.5\color{black}} \\
 &  Cross entropy  & \textbf{\color{red} 55.0\color{black}} &  \textbf{\color{red} 48.9\color{black}}  &  60.5 & \textbf{88.7} &  \textbf{\color{red} 85.5\color{black}} &  \textbf{91.0} &  \textbf{87.7} &  \textbf{\color{red}84.0\color{black}} &  \textbf{90.1} \\
\bottomrule
\end{tabular}
\label{performancelossfunctions}
\end{table*}

\begin{table*}
\caption{\color{black} Performance statistics in testing sets for two different sn\_ratio values, including mean, minimum, and maximum values for F$_1$, ROC-OVO, and ROC-OVA. Thirty experiments are summarised in each row, and 40,000 stars were used to train each model. Focal loss was used in these experiments. The row labelled ``Two losses" indicates that the training was conducted using synthetic light curves. In contrast, the row labelled "One loss" represents results from training with the same representation and architecture but without using synthetic samples.}
\begin{tabular}{llrrrrrrrrrr}
\toprule
{} & & sn\_ratio & \multicolumn{3}{c}{F$_1$} & \multicolumn{3}{c}{ROC-OVO} & \multicolumn{3}{c}{ROC-OVA} \\
&  &  &    mean &   min &   max &         mean &   min &   max &         mean &   min &   max \\
\midrule
 \textit{Data set I} & One loss  &        4 &    59.7 &  51.3 &  64.3 &         90.0 &  86.6 &  91.8 &         89.3 &  86.0 &  90.8 \\

&     &        6 &     60.7 &  52.3 &  66.1 &         90.4 &  87.7 &  91.6 &         89.3 &  87.3 &  90.8 \\
 &     &           8 &    60.4 &  48.6 &  66.7 &         90.4 &  87.1 &  92.1 &         89.4 &  85.5 &  91.0 \\
&     &          10 &    59.9 &  51.0 &  65.0 &         90.3 &  87.3 &  91.6 &         89.2 &  86.1 &  90.6 \\
 &  Two losses &        4 &    $^{\ast+}$61.2 &  55.6 &  67.0 &         $^{\ast+}$90.9 &  88.8 &  92.3 &         $^{\ast+}$90.2 &  87.6 &  92.1 \\
 &   &        6 &    60.8 &  55.7 &  67.0 &         90.6 &  87.2 &  91.7 &         $^{\ast+}$90.1 &  87.7 &  91.3 \\
 &   &        8 &    60.4 &  53.9 &  66.0 &         90.5 &  88.6 &  92.1 &         $^{\ast+}$90.0 &  87.5 &  91.5 \\
 &   &       10 &    60.5 &  55.6 &  65.7 &         90.5 &  87.4 &  92.3 &         $^{\ast+}$89.9 &  86.5 &  91.9 \\
\midrule
  \textit{Data set II} & One loss &  4 &  61.3 &  54.1 &  65.3 &  90.1 &  88.1 &  91.7 &  90.8 &  88.5 &  92.4 \\
 &     &        6 &    61.1 &  55.2 &  68.0 &         90.1 &  86.9 &  92.4 &         90.8 &  87.9 &  92.7 \\
 &     &        8 &    60.5 &  49.6 &  66.4 &         90.0 &  84.6 &  91.8 &         90.7 &  85.4 &  92.4 \\
&     &       10 &    60.2 &  51.6 &  65.3 &         89.9 &  86.6 &  91.2 &         90.6 &  87.9 &  91.9 \\
 &   Two losses &        4 &    61.5 &  36.8 &  67.4 &         $^{\ast}$90.3 &  78.6 &  92.3 &         $^{\ast+}$91.4 &  80.5 &  93.2 \\
 &    &        6 &    $^{\ast}$61.7 &  45.1 &  68.0 &         $^{\ast+}$90.6 &  82.6 &  92.2 &         $^{\ast+}$91.6 &  83.8 &  93.1 \\
 &    &        8 &    59.7 &  51.2 &  66.6 &         90.0 &  84.7 &  92.6 &         $^{\ast}$91.1 &  85.0 &  93.5 \\
 &    &       10 &    $^{\ast+}$61.4 &  46.7 &  66.3 &         $^{\ast+}$90.5 &  86.9 &  92.0 &         $^{\ast+}$91.6 &  88.1 &  93.0 \\

\bottomrule
\label{sn_ration_an}
\end{tabular}
\end{table*}
\color{black}

\begin{table*}
\caption{ \color{black} Performance metrics in testing sets for two different sequence lengths. Thirty experiments were conducted for each row using 40,000 stars for training.}
\begin{tabular}{lllrrrrrrrrrr}
\toprule
& {} & & seq\_length & \multicolumn{3}{c}{F$_1$} & \multicolumn{3}{c}{ROC-OVO} & \multicolumn{3}{c}{ROC-OVA} \\
& {} & &  &    mean &   min &   max &         mean &   min &   max &         mean &   min &   max \\
\midrule
 \textit{Data set I}& &    One loss &   50 &    46.4 &  38.1 &  53.2 &         84.7 &  79.7 &  86.3 &         82.2 &  76.5 &  84.2 \\
& &     &         150 &    57.4 &  51.1 &  61.6 &         88.8 &  87.3 &  90.6 &         87.5 &  85.5 &  89.2 \\
& &   Two losses &   50 &    $^{\ast+}$49.3 &  43.8 &  55.5 &         $^{\ast+}$85.6 &  81.3 &  87.2 &         $^{\ast+}$83.3 &  78.4 &  85.2 \\
& &   &        150 &    57.0 &  51.7 &  60.7 &         88.8 &  85.6 &  90.0 &         87.3 &  85.1 &  88.4 \\
\midrule
 \textit{Data set II}& &    One loss &         50 &    52.9 &  48.0 &  56.5 &         85.9 &  83.8 &  87.1 &         87.2 &  85.2 &  88.4 \\
& &    &     150 &    58.3 &  51.6 &  63.0 &         88.6 &  86.7 &  90.0 &         89.5 &  87.4 &  90.7 \\
& &   Two losses  &         50 &    53.2 &  49.3 &  57.3 &         85.6 &  82.1 &  87.4 &         86.9 &  83.8 &  88.5 \\
& &   &     150 &    58.0 &  49.7 &  62.8 &         88.5 &  86.1 &  90.1 &         89.5 &  87.2 &  90.9 \\
\bottomrule
\end{tabular}
\label{seqlength}
\end{table*}

\clearpage
\section{Conclusions}
\label{conclusion}

We introduced a novel approach to improving the reliability of variable star classifiers by handling the challenges posed by data shift and class imbalance. Our methodology combines a self-regulated CNN with a PELS-VAE to dynamically generate synthetic light curves that mitigate the underlying issues in data. This mechanism aims to train a classifier that generalises more effectively in unseen data.

The self-regulated CNN architecture, a key component of our methodology, utilises dual mask training to  distinguish between real and synthetic data patterns. The PELS-VAE model, conditioned on physical parameters, plays a crucial role by effectively generating realistic and highly informative synthetic light curves. These synthetic samples are generated from less probable physical parameter zones according to the classification performance obtained from the confusion matrix in each epoch.

The experiments prove that our approach improves the classification reliability and allows us to enhance the hyperparameter optimisation. In addition, our proposal evidences a stable behaviour under different conditions such as signal-to-noise ratio, loss functions and sequence length.   

Certain limitations in the current scenario constrain this approach's performance. Uncertainties in some physical parameters, such as metallicity, may hinder the PELS-VAE fitting process, thereby limiting the exploration of the physical space when generating new synthetic samples. Future technological and theoretical advances in physical parameter estimation should mitigate this issue. Additionally, systems could incorporate spectroscopy estimates for some physical parameters according to PELS-VAE requirements, using an active learning approach when deploying this model in real environments.

We propose five policies to manage the interaction between the classifier and the generative model. \color{black} These policies are based on greedy criteria, but other criteria could be proposed, or they could be used as part of a combined strategy. \color{black} Future contributions can focus on searching for optimal policies for each training set, and this policy can be learned during training. Moreover, these policies could be dynamic during the training phase. Certainly, this learning process can be very time-consuming, but it represents an interesting avenue for further exploration.

\section*{Acknowledgements}

We acknowledge the support from CONICYT-Chile,
through the FONDECYT Regular project number 1180054. F.P. acknowledges the support from National Agency for Research and Development (ANID), through Scholarship Program/Doctorado Nacional/2017-21171036. Support for M.C. is provided by ANID's FONDECYT Regular grant \#1171273; ANID's Millennium Science Initiative through grants ICN12\textunderscore 009 and AIM23-0001, awarded to the Millennium Institute of Astrophysics (MAS); and ANID's Basal project FB210003.

\section*{Data Availability}

The data underlying this article will be shared on reasonable request with the corresponding author. The code is available at \href{https://github.com/frperezgalarce/cnn-pels-vae}{https://github.com/frperezgalarce/cnn-pels-vae}. Data sets are available in \href{https://zenodo.org/records/13284675}{https://zenodo.org/records/13284675}.



\begin{thebibliography}{}
\makeatletter
\relax
\def\mn@urlcharsother{\let\do\@makeother \do\$\do\&\do\#\do\^\do\_\do\%\do\~}
\def\mn@doi{\begingroup\mn@urlcharsother \@ifnextchar [ {\mn@doi@} {\mn@doi@[]}}
\def\mn@doi@[#1]#2{\def\@tempa{#1}\ifx\@tempa\@empty \href {http://dx.doi.org/#2} {doi:#2}\else \href {http://dx.doi.org/#2} {#1}\fi \endgroup}
\def\mn@eprint#1#2{\mn@eprint@#1:#2::\@nil}
\def\mn@eprint@arXiv#1{\href {http://arxiv.org/abs/#1} {{\tt arXiv:#1}}}
\def\mn@eprint@dblp#1{\href {http://dblp.uni-trier.de/rec/bibtex/#1.xml} {dblp:#1}}
\def\mn@eprint@#1:#2:#3:#4\@nil{\def\@tempa {#1}\def\@tempb {#2}\def\@tempc {#3}\ifx \@tempc \@empty \let \@tempc \@tempb \let \@tempb \@tempa \fi \ifx \@tempb \@empty \def\@tempb {arXiv}\fi \@ifundefined {mn@eprint@\@tempb}{\@tempb:\@tempc}{\expandafter \expandafter \csname mn@eprint@\@tempb\endcsname \expandafter{\@tempc}}}

\bibitem[\protect\citeauthoryear{Abdollahi, Torabi, Raeisi  \& Rahvar}{Abdollahi et~al.}{2023}]{abdollahi2023hierarchical}
Abdollahi M.,  Torabi N.,  Raeisi S.,   Rahvar S.,  2023, Iranian Journal of Astronomy and Astrophysics, 2

\bibitem[\protect\citeauthoryear{Aguirre, Pichara  \& Becker}{Aguirre et~al.}{2019}]{aguirre2018deep}
Aguirre C.,  Pichara K.,   Becker I.,  2019, Monthly Notices of the Royal Astronomical Society, 482, 5078

\bibitem[\protect\citeauthoryear{Alcock et~al.,}{Alcock et~al.}{1997}]{alcock1997macho}
Alcock C.,  et~al., 1997, The Astrophysical Journal, 486, 697

\bibitem[\protect\citeauthoryear{Armstrong, G{\'o}mez Maqueo~Chew, Faedi  \& Pollacco}{Armstrong et~al.}{2013}]{armstrong2013catalogue}
Armstrong D.~J.,  G{\'o}mez Maqueo~Chew Y.,  Faedi F.,   Pollacco D.,  2013, Monthly Notices of the Royal Astronomical Society, 437, 3473

\bibitem[\protect\citeauthoryear{Auvergne et~al.,}{Auvergne et~al.}{2009}]{auvergne2009corot}
Auvergne M.,  et~al., 2009, Astronomy \& Astrophysics, 506, 411

\bibitem[\protect\citeauthoryear{Bassi, Sharma  \& Gomekar}{Bassi et~al.}{2021}]{bassi2021classification}
Bassi S.,  Sharma K.,   Gomekar A.,  2021, Frontiers in Astronomy and Space Sciences, 8, 168

\bibitem[\protect\citeauthoryear{Beaton et~al.,}{Beaton et~al.}{2016}]{beaton2016carnegie}
Beaton R.~L.,  et~al., 2016, The Astrophysical Journal, 832, 210

\bibitem[\protect\citeauthoryear{Becker, Pichara, Catelan, Protopapas, Aguirre  \& Nikzat}{Becker et~al.}{2020}]{becker2020scalable}
Becker I.,  Pichara K.,  Catelan M.,  Protopapas P.,  Aguirre C.,   Nikzat F.,  2020, Monthly Notices of the Royal Astronomical Society, 493, 2981

\bibitem[\protect\citeauthoryear{Benavente, Protopapas  \& Pichara}{Benavente et~al.}{2017}]{benavente2017automatic}
Benavente P.,  Protopapas P.,   Pichara K.,  2017, The Astrophysical Journal, 845, 147

\bibitem[\protect\citeauthoryear{Biewald}{Biewald}{2020}]{wandb}
Biewald L.,  2020, Experiment Tracking with Weights and Biases, \url {https://www.wandb.com/}

\bibitem[\protect\citeauthoryear{Blei \& Jordan}{Blei \& Jordan}{2006}]{blei2006variational}
Blei D.~M.,  Jordan M.~I.,  2006, Bayesian Analysis, 1, 121

\bibitem[\protect\citeauthoryear{Burgess, Higgins, Pal, Matthey, Watters, Desjardins  \& Lerchner}{Burgess et~al.}{2018}]{burgess2018understanding}
Burgess C.~P.,  Higgins I.,  Pal A.,  Matthey L.,  Watters N.,  Desjardins G.,   Lerchner A.,  2018, arXiv preprint arXiv:1804.03599

\bibitem[\protect\citeauthoryear{Burhanudin et~al.,}{Burhanudin et~al.}{2021}]{burhanudin2021light}
Burhanudin U.,  et~al., 2021, Monthly Notices of the Royal Astronomical Society, 505, 4345

\bibitem[\protect\citeauthoryear{Cabrera, Miller  \& Schneider}{Cabrera et~al.}{2014}]{cabrera2014systematic}
Cabrera G.~F.,  Miller C.~J.,   Schneider J.,  2014, in International Conference on Pattern Recognition. pp 4417--4422

\bibitem[\protect\citeauthoryear{Carrasco-Davis et~al.,}{Carrasco-Davis et~al.}{2019}]{carrasco2018deep}
Carrasco-Davis R.,  et~al., 2019, The Astronomical Society of the Pacific, 131, 108006

\bibitem[\protect\citeauthoryear{Castro, Protopapas  \& Pichara}{Castro et~al.}{2017}]{castro2017uncertain}
Castro N.,  Protopapas P.,   Pichara K.,  2017, The Astronomical Journal, 155, 16

\bibitem[\protect\citeauthoryear{Catelan \& Smith}{Catelan \& Smith}{2015}]{catelan2014pulsating}
Catelan M.,  Smith H.,  2015, Pulsating Stars.
Wiley-VCH, Weinheim

\bibitem[\protect\citeauthoryear{Copperwheat et~al.,}{Copperwheat et~al.}{2011}]{copperwheat2011sdss}
Copperwheat C.,  et~al., 2011, Monthly Notices of the Royal Astronomical Society, 410, 1113

\bibitem[\protect\citeauthoryear{Creevey et~al.,}{Creevey et~al.}{2023}]{creevey2023gaia}
Creevey O.~L.,  et~al., 2023, Astronomy \& Astrophysics, 674, A26

\bibitem[\protect\citeauthoryear{D{\'a}lya et~al.,}{D{\'a}lya et~al.}{2018}]{dalya2018glade}
D{\'a}lya G.,  et~al., 2018, Monthly Notices of the Royal Astronomical Society, 479, 2374

\bibitem[\protect\citeauthoryear{Debosscher et~al.,}{Debosscher et~al.}{2009}]{debosscher2009automated}
Debosscher J.,  et~al., 2009, Astronomy \& Astrophysics, 506, 519

\bibitem[\protect\citeauthoryear{Ding, Song, Wang  \& Ji}{Ding et~al.}{2024}]{ding2024detection}
Ding X.,  Song Z.,  Wang C.,   Ji K.,  2024, The Astronomical Journal, 167, 192

\bibitem[\protect\citeauthoryear{Donoso-Oliva, Becker, Protopapas, Cabrera-Vives, Vishnu  \& Vardhan}{Donoso-Oliva et~al.}{2023}]{donoso2022astromer}
Donoso-Oliva C.,  Becker I.,  Protopapas P.,  Cabrera-Vives G.,  Vishnu M.,   Vardhan H.,  2023, Astronomy \& Astrophysics, 670, A54

\bibitem[\protect\citeauthoryear{Drake et~al.,}{Drake et~al.}{2009}]{drake2009first}
Drake A.,  et~al., 2009, The Astrophysical Journal, 696, 870

\bibitem[\protect\citeauthoryear{Eyer et~al.,}{Eyer et~al.}{2023}]{eyer2023gaia}
Eyer L.,  et~al., 2023, Astronomy \& Astrophysics, 674, A13

\bibitem[\protect\citeauthoryear{Feast}{Feast}{1996}]{feast1996pulsation}
Feast M.,  1996, Monthly Notices of the Royal Astronomical Society, 278, 11

\bibitem[\protect\citeauthoryear{F{\"o}rster et~al.,}{F{\"o}rster et~al.}{2016}]{forster2016high}
F{\"o}rster F.,  et~al., 2016, The Astrophysical Journal, 832, 155

\bibitem[\protect\citeauthoryear{Garc{\'\i}a-Jara, Protopapas  \& Est{\'e}vez}{Garc{\'\i}a-Jara et~al.}{2022}]{garcia2022improving}
Garc{\'\i}a-Jara G.,  Protopapas P.,   Est{\'e}vez P.~A.,  2022, The Astrophysical Journal, 935, 23

\bibitem[\protect\citeauthoryear{Glorot \& Bengio}{Glorot \& Bengio}{2010}]{glorot2010understanding}
Glorot X.,  Bengio Y.,  2010, in Proceedings of the thirteenth international conference on artificial intelligence and statistics. pp 249--256

\bibitem[\protect\citeauthoryear{Goodfellow, Bengio  \& Courville}{Goodfellow et~al.}{2016}]{goodfellow2016deep}
Goodfellow I.,  Bengio Y.,   Courville A.,  2016, Deep learning.
The MIT Press, Cambridge, Massachusetts

\bibitem[\protect\citeauthoryear{Groenewegen}{Groenewegen}{2020}]{groenewegen2020analysing}
Groenewegen M.,  2020, Astronomy \& Astrophysics, 635, A33

\bibitem[\protect\citeauthoryear{Handler}{Handler}{2009}]{handler2009delta}
Handler G.,  2009, in AIP Conference Proceedings. pp 403--409

\bibitem[\protect\citeauthoryear{Higgins, Matthey, Pal, Burgess, Glorot, Botvinick, Mohamed  \& Lerchner}{Higgins et~al.}{2017}]{higgins2017beta}
Higgins I.,  Matthey L.,  Pal A.,  Burgess C.~P.,  Glorot X.,  Botvinick M.~M.,  Mohamed S.,   Lerchner A.,  2017, International Conference on Learning Representations (Poster), 3

\bibitem[\protect\citeauthoryear{Hosenie, Lyon, Stappers, Mootoovaloo  \& McBride}{Hosenie et~al.}{2020}]{hosenie2020imbalance}
Hosenie Z.,  Lyon R.,  Stappers B.,  Mootoovaloo A.,   McBride V.,  2020, Monthly Notices of the Royal Astronomical Society, 493, 6050

\bibitem[\protect\citeauthoryear{Ivezi{\'c} et~al.,}{Ivezi{\'c} et~al.}{2008}]{ivezic2008large}
Ivezi{\'c} {\v{Z}}.,  et~al., 2008, Serbian Astronomical Journal, pp 1--13

\bibitem[\protect\citeauthoryear{Jamal \& Bloom}{Jamal \& Bloom}{2020}]{jamal2020neural}
Jamal S.,  Bloom J.~S.,  2020, The Astrophysical Journal Supplement Series, 250, 30

\bibitem[\protect\citeauthoryear{Jayasinghe et~al.,}{Jayasinghe et~al.}{2019}]{jayasinghe2019asas}
Jayasinghe T.,  et~al., 2019, Monthly Notices of the Royal Astronomical Society, 486, 1907

\bibitem[\protect\citeauthoryear{Jayasinghe et~al.,}{Jayasinghe et~al.}{2021}]{jayasinghe2021asas}
Jayasinghe T.,  et~al., 2021, Monthly Notices of the Royal Astronomical Society, 503, 200

\bibitem[\protect\citeauthoryear{Jurcsik}{Jurcsik}{1995}]{jurcsik1995revision}
Jurcsik J.,  1995, Acta Astronomica, 45, 653

\bibitem[\protect\citeauthoryear{Kallrath, Milone  \& Wilson}{Kallrath et~al.}{2009}]{kallrath2009eclipsing}
Kallrath J.,  Milone E.~F.,   Wilson R.,  2009, Eclipsing binary stars: modeling and analysis.
~ Vol. 11, Springer

\bibitem[\protect\citeauthoryear{Kang, Zhang, Zhang, Li, Kong, Zhao  \& Wu}{Kang et~al.}{2023}]{kang2023periodic}
Kang Z.,  Zhang Y.,  Zhang J.,  Li C.,  Kong M.,  Zhao Y.,   Wu X.-B.,  2023, Publications of the Astronomical Society of the Pacific, 135, 094501

\bibitem[\protect\citeauthoryear{Kingma \& Welling}{Kingma \& Welling}{2013}]{kingma2013auto}
Kingma D.~P.,  Welling M.,  2013, Computing Research Repository in arXiv, abs/1312.6114

\bibitem[\protect\citeauthoryear{{Kolenberg, K.}, {Fossati, L.}, {Shulyak, D.}, {Pikall, H.}, {Barnes, T. G.}, {Kochukhov, O.}  \& {Tsymbal, V.}}{{Kolenberg, K.} et~al.}{2010}]{Kolenberg}
{Kolenberg, K.} {Fossati, L.} {Shulyak, D.} {Pikall, H.} {Barnes, T. G.} {Kochukhov, O.}  {Tsymbal, V.} 2010, \mn@doi [Astronomy \& Astrophysics] {10.1051/0004-6361/201014471}, 519, A64

\bibitem[\protect\citeauthoryear{Kovtyukh et~al.,}{Kovtyukh et~al.}{2023}]{kovtyukh2023effective}
Kovtyukh V.,  et~al., 2023, Monthly Notices of the Royal Astronomical Society, 523, 5047

\bibitem[\protect\citeauthoryear{Lin, Goyal, Girshick, He  \& Doll{\'a}r}{Lin et~al.}{2017}]{lin2017focal}
Lin T.-Y.,  Goyal P.,  Girshick R.,  He K.,   Doll{\'a}r P.,  2017, in Proceedings of the IEEE international conference on computer vision. pp 2980--2988

\bibitem[\protect\citeauthoryear{Lucas, Tucker, Grosse  \& Norouzi}{Lucas et~al.}{2019}]{lucas2019understanding}
Lucas J.,  Tucker G.,  Grosse R.,   Norouzi M.,  2019, in International Conference on Learning Representations. pp 1--16

\bibitem[\protect\citeauthoryear{MacFarland, Yates, MacFarland  \& Yates}{MacFarland et~al.}{2016}]{macfarland2016mann}
MacFarland T.~W.,  Yates J.~M.,  MacFarland T.~W.,   Yates J.~M.,  2016, Introduction to nonparametric statistics for the biological sciences using R, pp 103--132

\bibitem[\protect\citeauthoryear{Mahabal, Sheth, Gieseke, Pai, Djorgovski, Drake  \& Graham}{Mahabal et~al.}{2017}]{mahabal2017deep}
Mahabal A.,  Sheth K.,  Gieseke F.,  Pai A.,  Djorgovski S.,  Drake A.,   Graham M.,  2017, in Computational Intelligence, IEEE Symposium Series on. pp~1--8

\bibitem[\protect\citeauthoryear{Marconi, Nordgren, Bono, Schnider  \& Caputo}{Marconi et~al.}{2005}]{marconi2005predicted}
Marconi M.,  Nordgren T.,  Bono G.,  Schnider G.,   Caputo F.,  2005, The Astrophysical Journal, 623, L133

\bibitem[\protect\citeauthoryear{Mart{\'\i}nez-Palomera, Bloom  \& Abrahams}{Mart{\'\i}nez-Palomera et~al.}{2022}]{martinez2022deep}
Mart{\'\i}nez-Palomera J.,  Bloom J.~S.,   Abrahams E.~S.,  2022, The Astronomical Journal, 164, 263

\bibitem[\protect\citeauthoryear{Minniti et~al.,}{Minniti et~al.}{2010}]{minniti2010vista}
Minniti D.,  et~al., 2010, New Astronomy, 15, 433

\bibitem[\protect\citeauthoryear{Naul, Bloom, P{\'e}rez  \& van~der Walt}{Naul et~al.}{2018}]{naul2018recurrent}
Naul B.,  Bloom J.~S.,  P{\'e}rez F.,   van~der Walt S.,  2018, Nature Astronomy, 2, 151

\bibitem[\protect\citeauthoryear{Nun, Protopapas, Sim, Zhu, Dave, Castro  \& Pichara}{Nun et~al.}{2015}]{nun2015fats}
Nun I.,  Protopapas P.,  Sim B.,  Zhu M.,  Dave R.,  Castro N.,   Pichara K.,  2015, arXiv preprint arXiv:1506.00010

\bibitem[\protect\citeauthoryear{Pedregosa et~al.,}{Pedregosa et~al.}{2011}]{pedregosa2011scikit}
Pedregosa F.,  et~al., 2011, the Journal of machine Learning research, 12, 2825

\bibitem[\protect\citeauthoryear{P{\'e}rez-Galarce, Pichara, Huijse, Catelan  \& Mery}{P{\'e}rez-Galarce et~al.}{2021}]{perez2021informative}
P{\'e}rez-Galarce F.,  Pichara K.,  Huijse P.,  Catelan M.,   Mery D.,  2021, Monthly Notices of the Royal Astronomical Society, 503, 484

\bibitem[\protect\citeauthoryear{P{\'e}rez-Galarce, Pichara, Huijse, Catelan  \& Mery}{P{\'e}rez-Galarce et~al.}{2023}]{perez2023informative}
P{\'e}rez-Galarce F.,  Pichara K.,  Huijse P.,  Catelan M.,   Mery D.,  2023, Astronomy and Computing, 43, 100694

\bibitem[\protect\citeauthoryear{Pr{\v{s}}a \& Zwitter}{Pr{\v{s}}a \& Zwitter}{2005}]{prvsa2005computational}
Pr{\v{s}}a A.,  Zwitter T.,  2005, The Astrophysical Journal, 628, 426

\bibitem[\protect\citeauthoryear{Quiñonero-Candela, Sugiyama, Schwaighofer  \& Lawrence}{Quiñonero-Candela et~al.}{2009}]{candela2009dataset}
Quiñonero-Candela J.,  Sugiyama M.,  Schwaighofer A.,   Lawrence N.,  2009, Dataset shift in machine learning.
The MIT Press, Cambridge, Massachusetts

\bibitem[\protect\citeauthoryear{Richards}{Richards}{2012}]{richards2012overcoming}
Richards J.,  2012, in , Astrostatistics and Data Mining.
Springer, pp 213--221

\bibitem[\protect\citeauthoryear{Richards \& Groener}{Richards \& Groener}{2022}]{richards2022conditional}
Richards R.~J.,  Groener A.~M.,  2022, arXiv preprint arXiv:2205.01592

\bibitem[\protect\citeauthoryear{Richards et~al.,}{Richards et~al.}{2011}]{richards2011active}
Richards J.,  et~al., 2011, The Astrophysical Journal, 744, 192

\bibitem[\protect\citeauthoryear{Sesar et~al.,}{Sesar et~al.}{2013}]{sesar2013exploring}
Sesar B.,  et~al., 2013, The Astronomical Journal, 146, 21

\bibitem[\protect\citeauthoryear{Steeghs et~al.,}{Steeghs et~al.}{2022}]{steeghs2022gravitational}
Steeghs D.,  et~al., 2022, Monthly Notices of the Royal Astronomical Society, 511, 2405

\bibitem[\protect\citeauthoryear{Tisserand et~al.,}{Tisserand et~al.}{2007}]{tisserand2007limits}
Tisserand P.,  et~al., 2007, Astronomy \& Astrophysics, 469, 387

\bibitem[\protect\citeauthoryear{Trabucchi, Mowlavi  \& Lebzelter}{Trabucchi et~al.}{2021}]{trabucchi2021semi}
Trabucchi M.,  Mowlavi N.,   Lebzelter T.,  2021, Astronomy \& Astrophysics, 656, A66

\bibitem[\protect\citeauthoryear{Troyanskaya, Cantor, Sherlock, Brown, Hastie, Tibshirani, Botstein  \& Altman}{Troyanskaya et~al.}{2001}]{troyanskaya2001missing}
Troyanskaya O.,  Cantor M.,  Sherlock G.,  Brown P.,  Hastie T.,  Tibshirani R.,  Botstein D.,   Altman R.~B.,  2001, Bioinformatics, 17, 520

\bibitem[\protect\citeauthoryear{Tsang \& Schultz}{Tsang \& Schultz}{2019}]{tsang2019deep}
Tsang B. T.-H.,  Schultz W.~C.,  2019, The Astrophysical Journal Letters, 877, L14

\bibitem[\protect\citeauthoryear{Udalski, Szymanski, Soszynski  \& Poleski}{Udalski et~al.}{2008}]{udalski2008optical}
Udalski A.,  Szymanski M.~K.,  Soszynski I.,   Poleski R.,  2008, Acta Astron., 58, 69

\bibitem[\protect\citeauthoryear{Udalski, Szyma{\'n}ski  \& Szyma{\'n}ski}{Udalski et~al.}{2015}]{udalski2015ogle}
Udalski A.,  Szyma{\'n}ski M.,   Szyma{\'n}ski G.,  2015, Acta Astronomica, 65

\bibitem[\protect\citeauthoryear{Uytterhoeven et~al.,}{Uytterhoeven et~al.}{2011}]{uytterhoeven2011kepler}
Uytterhoeven K.,  et~al., 2011, Astronomy \& Astrophysics, 534, A125

\bibitem[\protect\citeauthoryear{Vilalta, Gupta  \& Macri}{Vilalta et~al.}{2013}]{vilalta2013machine}
Vilalta R.,  Gupta K.~D.,   Macri L.,  2013, Astronomy and Computing, 2, 46

\bibitem[\protect\citeauthoryear{Wright et~al.,}{Wright et~al.}{2010}]{wright2010wide}
Wright E.~L.,  et~al., 2010, The Astronomical Journal, 140, 1868

\bibitem[\protect\citeauthoryear{Yang, Deb  \& Fong}{Yang et~al.}{2014}]{yang2014metaheuristic}
Yang X.-S.,  Deb S.,   Fong S.,  2014, Applied Mathematics \& Information Sciences, 8, 977

\bibitem[\protect\citeauthoryear{Zhang \& Bloom}{Zhang \& Bloom}{2021}]{zhang2021classification}
Zhang K.,  Bloom J.~S.,  2021, Monthly Notices of the Royal Astronomical Society, 505, 515

\makeatother
\end{thebibliography}





\appendix
\section{Supplementary material}

\begin{table}
\centering
\caption{Number $N$ of missing astrophysical parameters. Metallicity ([Fe/H]), effective temperature ($T_{\rm eff}$), period ($P$), absolute G magnitude ($M_G$), radius ($R$), and surface gravity ($\log g$).
}
\label{missingdata}
\begin{tabular}{lrrrrrrr}
\toprule
Type &   $N_{\rm[Fe/H]}$ &  $N_{T_{\rm eff}}$ &  $N_P$ &  $N_{M_G}$ &  $N_R$ &  $N_{\log g}$ \\
\midrule
CEP   &             1,735 &       339 &       0 &         0 &        1,734 &  1,735 \\
DSCT  &              634 &       288 &       0 &         0 &         634 &   634 \\
ECL   &             2,685 &       458 &       0 &         0 &        2,491 &  2,685 \\
LPV   &           6,331 &       702 &       0 &         0 &        6,326 &  6,331 \\
RRLYR &            4,667 &      3,709 &       0 &         0 &        5,065 &  5,076 \\
\bottomrule
\end{tabular}
\end{table}

\begin{table*}
\caption{Overview of approximate ranges of physical parameters for included variable stars in this study. }
\centering
\begin{tabular}{lllcccccc}
\toprule
Star Type & Complete Name &  $T_{\rm eff}$ & $P$ & $M_G$ & \( \log g \) & [Fe/H] & R  \\
 & &  (K) & (days) & (G band) & (dex) & (dex) & ($R_\odot$) \\

\midrule
RR Lyrae & RR Lyrae variables  & 6,200 - 6,800 & 0.30 - 0.90 & 0.30 to 1.00 & 2.5 - 3.5 & -2 to 0 & 4.0 - 6.5 \\
CEP      & Classical Cepheids     & 5,000 - 8,000 & 1 - 200 & -2 to +2 & 1.0 - 4.0 & -0.5 to 0.5 & 5 - 50 \\
DSCT     & Delta Scuti variables  & 6,500 - 8,500 & 0.02 - 0.3 & +0.5 to +4.0 & 3.5 - 4.5 & -0.5 to 0.5 & 1.5 - 2.5 \\
ECL      & Eclipsing binaries  & 2,500 - 30,000 & 0.1 - 10,000 & -3 to +10 & Varies  & Varies & Varies \\
LPV      & Long-period variables & 2,000 - 3,500 & 100 - 1,000 & -1.5 to +2.0 & 0 - 2.0 & -3 to 0 & 50 - 500 \\
\bottomrule
\end{tabular}
\begin{tablenotes}
        \footnotesize
        \item $^1$ References for physical parameters values \citep{feast1996pulsation,marconi2005predicted,handler2009delta,kallrath2009eclipsing,Kolenberg,uytterhoeven2011kepler,copperwheat2011sdss,armstrong2013catalogue,catelan2014pulsating,groenewegen2020analysing,jayasinghe2021asas,trabucchi2021semi,kovtyukh2023effective,eyer2023gaia}.
    \end{tablenotes}
\label{physicalparameters}
\end{table*}




\begin{figure*}
    \centering
    \begin{subfigure}[b]{1\textwidth}
        \includegraphics[width=\textwidth, trim=0 0 0 0, clip=True]{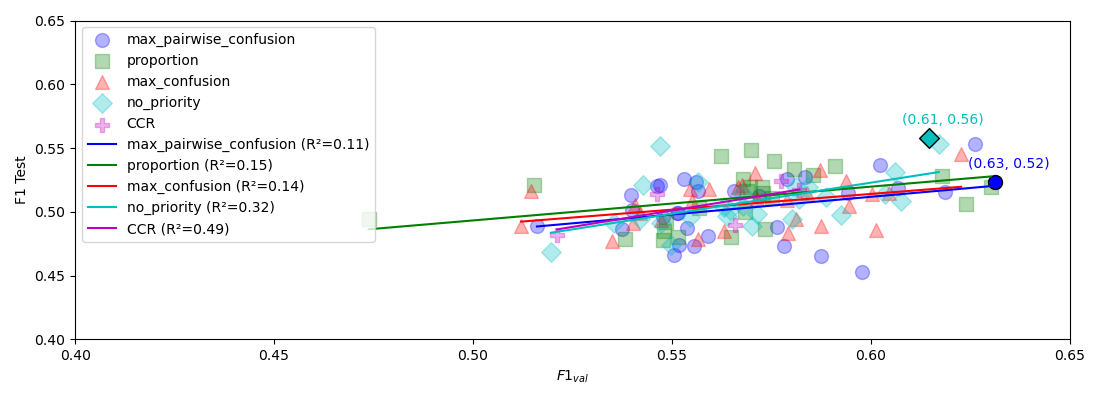}
        \caption{}
        \label{hyperparameterSearch1}
    \end{subfigure}
        \begin{subfigure}[b]{1\textwidth}
        \includegraphics[width=\textwidth,  trim=0 0 0 0, clip=True]{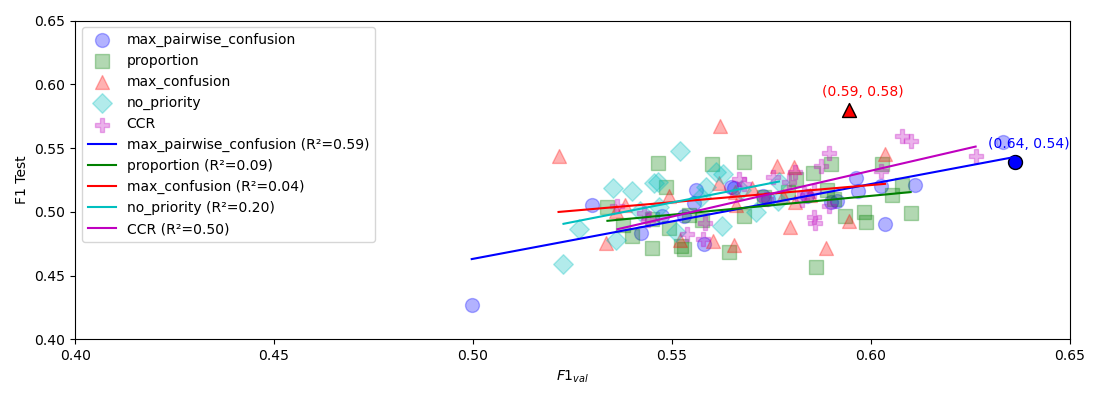}
        \caption{}
        \label{hyperparameterSearch3}
    \end{subfigure}
    \begin{subfigure}[b]{1\textwidth}
        \includegraphics[width=\textwidth,  trim=0 0 0 0, clip=True]{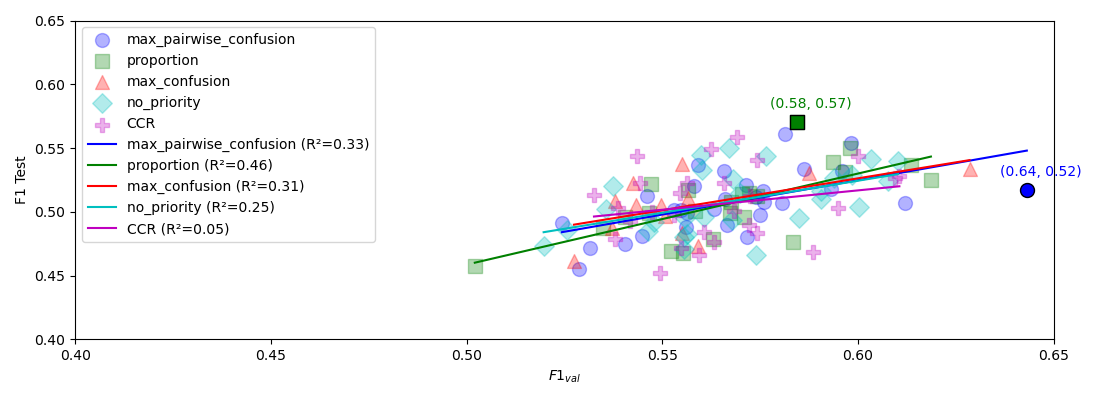}
        \caption{}
        \label{hyperparameterSearch2}
    \end{subfigure}
    \caption{Hyperparameter search results using Bayesian optimisation with \texttt{Weights \& Biases}. Points represent different configurations, colour-coded by the policy employed for defining the number of samples for each class. Regression lines show relationships between the objective score (F$_1^{\rm val}$ or F$_1^{\rm weighted}$) and F$_1$ test scores. 120 \color{black} loss function \color{black} evaluations were applied. (a) F$_1^{val}$ is used as objective score. (b) F$_1^{\rm weighted}$ is used as objective score, F$_1^{\rm weighted} = 0.85*$F$_1^{\rm val} + 0.15*$F$_1^{\rm synthetic}$ , where F$_1^{\rm val}$ is the macro weighted F$_1$ from the training set and F$_1^{\rm synthetic}$ is the macro weighted F$_1$ from synthetic samples. (c) F$_1^{\rm weighted}$ is used as objective score, F$_1^{\rm weighted}$ = 0.7*F$_1^{\rm val}$+ 0.3*F$_1^{\rm synthetic}$. Each figure, through its legend, shows the  coefficient of determination, \textbf{R}$^2$ of F$_1$ on the validation set and F$_1$ on the testing set. This metric indicates how effective the search is in terms of the correlation between the performances on the validation set and the test set. For four out of five policies, the best correlation was found in a setting that includes synthetic light curves during the hyperparameter search.}
    \label{fig:hyperparameter_search_combined}
\end{figure*}
\label{lastpage}
\end{document}